\documentclass[conference]{IEEEtran}
\IEEEoverridecommandlockouts % This "breaks" the conference rules to allow \thanks
\usepackage{times}

\usepackage[numbers,sort&compress]{natbib}
\usepackage{graphicx}

\usepackage{multicol}
\usepackage[bookmarks=true]{hyperref}
\usepackage{caption}  % For \captionof command
\usepackage{cuted}
\usepackage[table]{xcolor}
\usepackage{amsmath}
\usepackage{amssymb}
\usepackage{bm}

\usepackage{booktabs}  % for \toprule, \midrule, \bottomrule
\usepackage{multirow}  % for Option 4
\usepackage{pifont}    % for Option 1 (\cmark)

\begin{document}

% paper title
\title{Sim-to-Real of Humanoid Locomotion Policies via Joint Torque Space Perturbation Injection}

% You will get a Paper-ID when submitting a pdf file to the conference system
% \author{Author Names Omitted for Anonymous Review. Paper-ID [add your ID here]}

\author{Junhyeok Rui Cha$^{1}$, Woohyun Cha$^{1}$, Jaeyong Shin$^{1}$, Donghyeon Kim$^{1,2}$ and Jaeheung Park$^{3}$% <-this % stops a space
% \thanks{*This work was not supported by any organization}% <-this % stops a space
\thanks{$^{1}$Department of Intelligence and Information, Graduate School of Convergence Science and Technology, Seoul National University, Republic of Korea
\{{\tt\small threeman1, woohyun321, jasonshin0537, kdh0429}\}@snu.ac.kr}%
\thanks{$^{2}$1X Technologies. This work was conducted while the author was at Seoul National University}%
\thanks{$^{3}$Department of Intelligence and Information, Graduate School of Conver
gence Science and Technology, ASRI, AIIS, Seoul National University,
Republic of Korea, and Advanced Institute of Convergence Technology
(AICT), Suwon, Republic of Korea. He is the corresponding author of this paper. {\tt\small park73@snu.ac.kr}}%
}

\maketitle

\begin{figure*}[t]
    \centering
    \includegraphics[width=\textwidth]{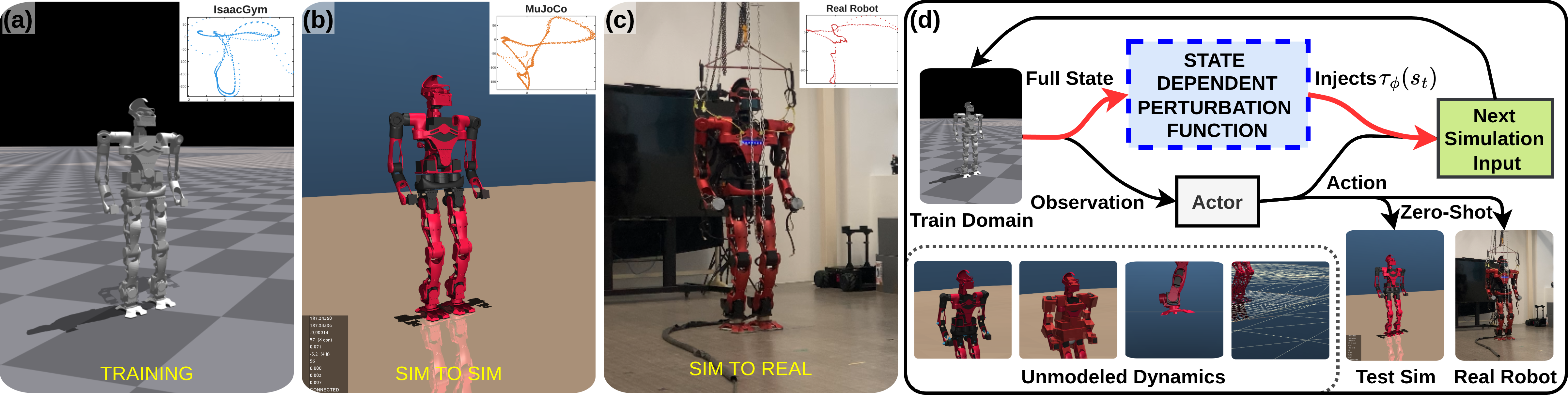}
    \captionof{figure}{\textbf{Snapshots of training, sim-to-sim transfer, and sim-to-real transfer.} This work proposes a novel sim-to-real method that injects state-dependent joint torque perturbations for enhanced robustness against complex reality gaps. Experimental results demonstrate that the proposed approach outperforms existing methods in both simulation and real-world environments.}
    \label{fig:isaac_torque_vel}
\end{figure*}

\begin{abstract}
This paper proposes a novel alternative to existing sim-to-real methods for training control policies with simulated experiences. Unlike prior methods that typically rely on domain randomization over a fixed finite set of parameters, the proposed approach injects state-dependent perturbations into the input joint torque during forward simulation. These perturbations are designed to simulate a broader spectrum of reality gaps than standard parameter randomization without requiring additional training. By using neural networks as flexible perturbation generators, the proposed method can represent complex, state-dependent uncertainties, such as nonlinear actuator dynamics and contact compliance, that parametric randomization cannot capture. Experimental results demonstrate that the proposed approach enables humanoid locomotion policies to achieve superior robustness against complex, unseen reality gaps in both simulation and real-world deployment.
\end{abstract}

\section{Introduction} \label{sec:Introduction}
Deep Reinforcement Learning (DRL) for robotic applications has gained significant attention due to its potential for robustness and versatility. Although DRL algorithms are capable of solving complex, high-dimensional control problems, commonly used on-policy methods often require a prohibitively large amount of data, posing a substantial challenge when collecting sufficient samples solely from real hardware. Furthermore, the exploration process required for policy improvement during early training stages can raise safety concerns for both the physical robot and its operational environment. Consequently, it has become standard practice to train control policies in simulation by creating a model (e.g., URDF) of the robot, configuring simulation parameters to approximate real-world conditions, and collecting rollouts to iteratively update the policies. Nonetheless, an unavoidable discrepancy---often referred to as the \textit{reality gap}---emerges between simulation and the real world, hindering direct policy deployment on physical systems.

To mitigate this gap, \textit{Domain Randomization} (DR) is widely adopted~\cite{tobin2017domain, peng2018sim}. This involves randomizing a fixed, finite set of simulation parameters during training, encouraging the resulting control policy to generalize over a broad range of domains, ideally including the real-world configuration. An effective deployment of DR typically requires: (1) constructing a nominal simulation environment that resembles the real system as closely as possible, and (2) selecting an optimal randomization range that balances policy generalization with performance~\cite{kumar2021rma}. Despite its success in numerous applications, DR is inherently constrained by the finite scope of parameters available for randomization. Moreover, many simulators either do not expose all necessary parameters or rely on dynamics formulations that lack sufficient expressiveness to capture full real-world complexities~\cite{11128640, pennestri2016review, bittencourt2012static}. As a result, control policies trained under these constraints often exhibit limited robustness to real-world variations.

To overcome these shortcomings, this work presents a novel sim-to-real framework that addresses the limitations of traditional DR by injecting state-dependent perturbations into the joint torque space. Motivated by the insight that the reality gap manifests primarily as variations in joint torque dynamics, the proposed method employs neural networks to introduce perturbations that are more expressive than those achievable through fixed-parameter randomization. Furthermore, by leveraging Denoising World Model Learning (DWL)~\cite{gu2024advancing}, the approach moves beyond strict stylistic imitation, enabling the learning of robust latent representations capable of handling diverse and challenging terrains. This method effectively captures complex, unmodeled dynamics---such as actuator discrepancies and contact compliance---without relying on extensive real-world data collection. Experimental validations on the full-sized humanoid robot TOCABI demonstrate that the framework achieves superior zero-shot sim-to-real transfer, maintaining high tracking performance even in the presence of significant reality gaps.

The contributions of this work are summarized as follows:
\begin{itemize}
\item \textbf{A state-dependent joint torque perturbation method for sim-to-real transfer.} This work first establishes that the effect of domain randomization can be equivalently viewed as perturbations in joint torque space, whose expressiveness is inherently limited by the simulator formulation. Based on this insight, the proposed method replaces parametric randomization with neural network-generated perturbations that depend on the robot state. This enables simulation of complex dynamics mismatches, such as nonlinear actuator behaviors and contact compliance, that fixed-parameter randomization cannot represent.
\item \textbf{Extensive validation on the full-sized humanoid robot TOCABI (180\,cm, 100\,kg).} Experiments demonstrate superior robustness against extended randomization ranges, unseen ground friction conditions, rough terrain, and modified dynamics, achieving successful zero-shot sim-to-real transfer for both position- and torque-based policies.
\item \textbf{A motion-capture-free training framework combining DWL with Raibert-based heuristics.} A GRU-based encoder learns robust latent representations via privileged observation reconstruction, while online Raibert-based trajectory generation provides velocity-adaptive references for reward shaping, enabling natural humanoid gaits without adversarial training.
\end{itemize}

% =====================
% RELATED WORK
% =====================
\section{Related Work} \label{sec:RelatedWork}

\subsection{DRL in Legged Robots} \label{subsec:DRLInLeggedRobotics}
Deep Reinforcement Learning (DRL) has shown significant promise in enhancing the robustness and versatility of legged robot control. Although some prior studies have employed off-policy DRL algorithms trained directly on physical platforms \cite{smith2022walk, haarnoja2018soft, wu2023daydreamer}, most research on legged robots relies on on-policy methods. Despite their high sample complexity, widely used on-policy algorithms offer advantages such as implementation simplicity and training stability. To address the large sample requirements, researchers often leverage high-fidelity simulators for data collection. The use of simulation also promotes safety in the early stages of training, preventing potential damage to real hardware caused by exploratory actions. However, policies trained exclusively in simulation do not readily transfer to physical systems due to discrepancies predominantly caused by actuator dynamics and communication latencies \cite{neunert2017off, ibarz2021train}. In addition, recent work has shown that inaccuracies in ground contact dynamics modeling significantly degrade the performance of locomotion policies \cite{choi2023learning}.

\subsection{Sim-to-Real Methods}\label{subsec:Sim2RealMethods}
A wide range of techniques have been explored to bridge the reality gap for successful policy deployment on physical robots. One straightforward approach is to calibrate the simulator to closely match the real-world robot. For example, \cite{tan2018sim} conducted rigorous system identification to replicate the dynamics of a quadrupedal platform in simulation, achieving successful sim-to-real transfer. Other approaches learn correction models by training neural networks on real hardware data to estimate and compensate for the reality gap \cite{jeong2019modelling, hwangbo2019learning}, thereby improving simulation fidelity. However, these methods often require extensive data collection, which may be infeasible when the hardware lacks certain sensing capabilities, such as link-side joint torque measurements \cite{hwangbo2019learning}, or when a partially functional policy is unavailable \cite{jeong2019modelling}.

Among sim-to-real techniques, Domain Randomization (DR) is the most widely adopted approach. DR involves randomizing selected simulation parameters related to dynamics \cite{peng2018sim}, robot models \cite{feng2023genloco, exarchos2021policy}, and observation noise or system latencies \cite{tan2018sim} during training. DR optimizes the control policy performance across these randomized domains; therefore, successful transfer ideally requires that the real-world system be encompassed within the training distribution \cite{kumar2021rma}. To this end, practitioners typically set nominal simulation parameters to match real-world conditions as closely as possible, based on system identification \cite{valassakis2020crossing} or manufacturer data \cite{kim2023torque}. Nonetheless, the diversity of environments generated via DR remains constrained by: (1) the finite set of parameters exposed by the simulator, and (2) the assumed formulations governing how these parameters affect the simulated physics. For example, simulators that provide limited access to contact dynamics or those that fail to accurately capture intricate actuator behaviors may yield control policies that fail under substantial real-world discrepancies. An alternative, though less prevalent, approach is \textit{Random Force Injection} (RFI) \cite{valassakis2020crossing, campanaro2024learning}, which introduces random biases and noise to the actuated components during forward simulation. By training the control policy in this stochastically perturbed environment, RFI achieves sim-to-real performance comparable to DR while reducing the complexity of parameter tuning.

\subsection{Learning Humanoid Locomotion}
The experiments and validations in this paper are conducted on TOCABI \cite{schwartz2022design}, a full-sized humanoid robot. Learning stable and natural locomotion for humanoids is particularly challenging due to the inherent instability of bipedal walking and the system high dimensionality, which allows for many asymmetric or unstable gait solutions \cite{yu2018symmetric}. Existing work generally addresses these issues through extensive reward shaping~\cite{duan2021rewardshaping, siekmann2021rewardshaping1, rudin2022rewardshaping2, radosavovic2024rewardshaping3} or motion imitation~\cite{peng2018deepmimic, peng2020deepmimichardware, peng2021amp, escontrela2022ampforhardware, tang2024humanmimic}.

While reward shaping demands domain expertise and significant engineering effort, motion imitation depends heavily on the quality and coverage of reference data. To bypass manual reward engineering, the Adversarial Motion Prior (AMP) framework~\cite{peng2021amp, escontrela2022ampforhardware, tang2024humanmimic} is widely adopted. However, as a Generative Adversarial Network-based method, AMP is susceptible to mode collapse and training instability. Crucially, it struggles with out-of-distribution (OOD) states; when reference datasets are limited, AMP often fails to generalize to rough terrain or generate necessary auxiliary behaviors, such as stable lateral movement.

In this work, DWL~\cite{gu2024advancing} is utilized to overcome these limitations. Unlike AMP, which relies on strict stylistic imitation that can hinder traversability on complex surfaces, DWL learns a robust latent representation using fixed rewards and state reconstruction. This approach enables the robot to learn stable locomotion policies capable of traversing rough terrain and performing lateral maneuvers, even where simple imitation-based policies fail.

\begin{figure}[t!]
    \centering
    \includegraphics[width=\linewidth]{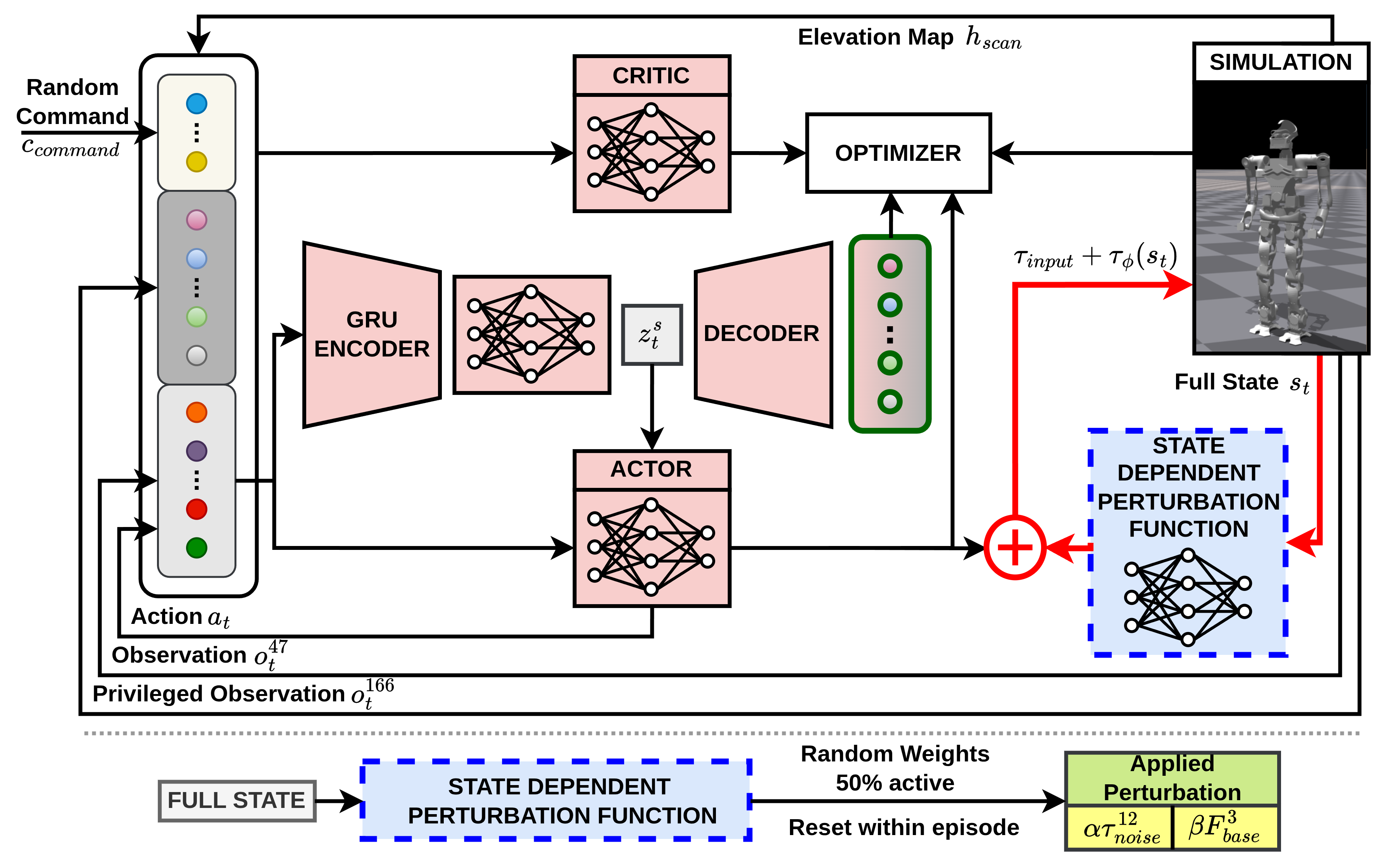}
    \caption{\textbf{Overview of the training framework: }The dynamics of the training environment is perturbed by the output of the randomly sampled state-dependent joint torque space perturbation function. The perturbation function is implemented as a neural network whose weights are randomly sampled at the beginning of each episode.  Details on deployment, DWL, and Raibert-based reference trajectories are provided in the Appendix \ref{app:deployment_raibert}.}
    \label{fig:training_framework}
\end{figure}

%=============================
\section{Method} \label{sec:Method}

\subsection{Reinforcement Learning}
The robot locomotion control problem is modeled as a discrete-time partially observable Markov Decision Process (POMDP). At each time step $t$, the agent makes an observation $o_t$ containing partial information about the system state $s_t$. Given $o_t$, the agent executes an action $a_t \sim \pi_\theta(\cdot|o_t)$ according to its control policy $\pi_\theta(\cdot|o)$. Following the execution of $a_t$, the environment transitions to its next state $s_{t+1} \sim P(\cdot|s_t, a_t)$ and the agent receives a reward $r_t = r(s_t, a_t)$. The objective is to maximize the expected cumulative discounted reward over a finite horizon $T$, defined as $J(\pi_\theta) = \mathbb{E}_{\tau \sim p(\tau|\pi_\theta)}[\sum_{t=0}^{T-1} \gamma^t r_t]$.

% --- Observation Space ---
\subsubsection{Observation Space} \label{subsubsec:observation_space}
The observation space is composed of the proprioceptive observation $o_t \in \mathcal{O} \subset \mathbb{R}^{47}$ available to the policy network, and the privileged observation $o_{\text{priv},t} \in \mathcal{O}_{\text{priv}}$, which contains ground-truth state information.

The proprioceptive observation $o_t$ consists of the base angular velocity $\omega_{\text{base},t} \in \mathbb{R}^3$, the gravity vector projected onto the robot body frame $g_{\text{proj},t} \in \mathbb{R}^3$, the command vector $c_{\text{cmd}} \in \mathbb{R}^3$, the deviation of joint positions from their default pose $q_t - q_{\text{default}} \in \mathbb{R}^{12}$, joint velocities $\dot{q}_t \in \mathbb{R}^{12}$, a gait phase signal $p_t \in \mathbb{R}^2$, and the previously executed action $a_{t-1} \in \mathbb{R}^{12}$:
\begin{equation}
o_t = [\omega_{\text{base},t}, g_{\text{proj},t}, c_{\text{cmd}}, q_t - q_{\text{default}}, \dot{q}_t, p_t, a_{t-1}]
\end{equation}

The gait phase signal $p_t$ consists of continuous sinusoidal encodings of the current gait cycle timing:
\begin{equation}
p_t = \left[ \cos\left(\frac{2\pi(t + \phi)}{2T}\right), \sin\left(\frac{2\pi(t + \phi)}{2T}\right) \right]
\end{equation}
where $t$ is the current step tick, $T$ is the step period, and $\phi$ is the phase indicator that distinguishes between legs.

The privileged observation $o_{\text{priv},t}$ augments $o_t$ with ground-truth states that are difficult to estimate accurately or unavailable on the real robot. This includes the base linear velocity $v_{\text{base},t} \in \mathbb{R}^3$, the relative position and yaw of the swing foot in the stance foot frame $x_{\text{swing}} \in \mathbb{R}^4$, the target swing foot state $x_{\text{swing}}^{\text{target}} \in \mathbb{R}^4$, the current reward $r_t$, binary foot contact states $c_{\text{contact}} \in \{0,1\}^2$, and the injected perturbation forces used in the proposed method:
\begin{equation}
o_{\text{priv},t} = [o_t, v_{\text{base},t}, x_{\text{swing}}, x_{\text{swing}}^{\text{target}}, r_t, c_{\text{contact}}, \tau_{\text{pert}}, F_{\text{body,pert}}, h_{\text{scan}}]
\end{equation}

Crucially, $o_{\text{priv},t}$ includes the explicit values of the injected joint torque perturbations $\tau_{\text{pert}}$ and body force perturbations $F_{\text{body,pert}}$. Access to these values allows the critic to correctly estimate the value function even in the presence of significant randomization introduced by the proposed method. Additionally, a height scan $h_{\text{scan}}$ of the surrounding terrain is included.

\subsubsection{Action Space}
The position- and torque-based control policy method extended from~\cite{kim2023torque} is used in this work.

The action space $\mathcal{A} \subset \mathbb{R}^{12}$ is composed of 12 actions, each used to generate the command for the corresponding lower body joint. The upper body joints are controlled to remain at their default positions using joint-space PD control. The actions $a_t$ are sampled at a frequency of 125\,Hz and stored in a history buffer to simulate system latency; the effective action $a_{\text{eff}}$~\cite{cha2023impact} is retrieved from this buffer with a configurable delay.

In the case of position-based control, the action represents normalized target joint positions bounded in the range $[-1, 1]$. These are mapped to the physical joint limits using the half-range $\sigma_{\text{pd}}$ and mean $\mu_{\text{pd}}$ of the limits, and the torque vector is computed via a PD law:
\begin{equation}
\tau_{\text{input},t} = K_p (\text{clip}(a_{\text{eff}}, -1, 1) \odot \sigma_{\text{pd}} + \mu_{\text{pd}} - q_t) - K_d \dot{q}_t
\end{equation}
where $K_p$ and $K_d$ are the proportional and derivative gain vectors, and $\odot$ denotes element-wise multiplication.

In the case of torque-based control, the action represents normalized torque commands bounded in the range $[-1, 1]$. The torque vector is computed by element-wise multiplication of the torque limits vector with the corresponding action vector:
\begin{equation}
\tau_{\text{input},t} = \tau_{\text{limit}} \odot \text{clip}(a_{\text{eff}}, -1, 1) \quad [\text{N}\cdot\text{m}]
\end{equation}

\subsubsection{Reward Function}
To guide the policy toward robust, natural, and command-responsive locomotion, the total reward $r_{\text{total}}$ is computed as a weighted sum of individual reward terms. These terms are categorized into three primary objectives: \textit{base command tracking}, which ensures precise adherence to user-defined velocity targets; \textit{motion quality}, which utilizes custom online heuristic references to encourage physically plausible gait patterns; and \textit{regularization}, which facilitates sim-to-real transfer by penalizing excessive energy consumption, unstable oscillations, heavy impacts, and jerky control actions. Detailed mathematical formulations and specific weights for all reward terms are provided in the Appendix \ref{app:rewards}.

\subsection{Domain Randomization as Joint Torque Space Perturbation}
\label{subsec:dr_torque_space}
This work focuses on overcoming the limitations of domain randomization methods for robot control. Consider the following dynamics equation used for forward simulation, perturbed by randomizing a fixed finite set of simulation parameters:
\begin{equation}
\begin{split}
M(q; p_{\text{DR}})\ddot{q} + C(q, \dot{q}; p_{\text{DR}}) + G(q; p_{\text{DR}}) + \tau_{\text{contact}}(s; p_{\text{DR}})\\ = \tau_{\text{output}}(\tau_{\text{input}}, s; p_{\text{DR}})
\end{split}
\end{equation}
where $s$ is the full state of the simulation dynamics. The terms in the forward simulation equation are conditioned on the simulation parameter set instance $p_{\text{DR}} \in \mathcal{P}_{\text{simul}}$, where $\mathcal{P}_{\text{simul}}$ is the set of all possible simulation configurations. Consider the effects of the perturbation on simulation parameters, formulated as:
\begin{align}
M(q; p_{\text{DR}})
&= \bar{M}(q) + \tilde{M}(q; p_{\text{DR}}) \nonumber \\
C(q, \dot{q}; p_{\text{DR}}) &= \bar{C}(q, \dot{q})
+ \tilde{C}(q, \dot{q}; p_{\text{DR}}) \nonumber \\
G(q; p_{\text{DR}})
&= \bar{G}(q) + \tilde{G}(q; p_{\text{DR}}) \nonumber \\
\tau_{\text{contact}}(s; p_{\text{DR}}) &= \bar{\tau}_{\text{contact}}(s) + \tilde{\tau}_{\text{contact}}(s; p_{\text{DR}}) \nonumber \\
\tau_{\text{output}}(\tau_{\text{input}}, s; p_{\text{DR}}) &= \tau_{\text{input}} + \tilde{\tau}_{\text{actuator}}(\tau_{\text{input}}, s; p_{\text{DR}})
\end{align}
where $\bar{\cdot}$ indicates values from unperturbed simulation parameters and $\tilde{\cdot}$ represents deviations between the unperturbed parameters and their perturbed counterparts. From the above equations, the following joint-space dynamics equation is obtained:
\begin{align}
\bar{M}(q)\ddot{q} + \bar{C}(q, \dot{q}) + \bar{G}(q) + \bar{\tau}_{\text{contact}}(s) &= \tau_{\text{input}} + \tau_{\text{DR}}(s; p_{\text{DR}})
\end{align}
where
\begin{equation}
\begin{split}
\tau_{\text{DR}}(s; p_{\text{DR}}) = -[\tilde{M}(q; p_{\text{DR}})\ddot{q} + \tilde{C}(q, \dot{q}; p_{\text{DR}}) + \tilde{G}(q; p_{\text{DR}})\\ + \tilde{\tau}_{\text{contact}}(s; p_{\text{DR}}) + \tilde{\tau}_{\text{actuator}}(\tau_{\text{input}}, s; p_{\text{DR}})]
\end{split}
\label{eq:tau_dr}
\end{equation}

This derivation establishes a key insight: \textit{the effects of randomized simulation parameters are equivalent to a nonlinear joint torque space perturbation function} $\tau_{\text{DR}}(\cdot; p_{\text{DR}})$, whose formulation is fixed by the simulator and physics engine. Define the set of all achievable perturbation functions as $\mathcal{F}_{\text{DR}}(\mathcal{P}_{\text{simul}}) = \{\tau_{\text{DR}}(\cdot; p_{\text{DR}}) \mid p_{\text{DR}} \in \mathcal{P}_{\text{simul}}\}$. Successful sim-to-real transfer via DR requires that $\tau_{\text{reality gap}}(\cdot) \in \mathcal{F}_{\text{DR}}(\mathcal{P}_{\text{simul}})$.

However, $\mathcal{F}_{\text{DR}}$ is inherently limited by: (1) restricted access to $\mathcal{P}_{\text{simul}}$ (e.g., contact softness parameters often cannot be modified mid-training), and (2) insufficient expressiveness of $\tau_{\text{DR}}$ (e.g., simplified actuator models that fail to capture complex dynamics such as Stribeck friction, harmonic drive hysteresis, or position-dependent cogging torque).

\subsection{Replacing Parametric Randomization with Neural Network Perturbations}
\label{subsec:nn_function_class}
The preceding analysis reveals a key limitation: DR can only produce perturbation functions within the constrained class $\mathcal{F}_{\text{DR}}$. To overcome this, the proposed method replaces parametric randomization with neural network-generated perturbations:
\begin{equation}
\tau_{\phi}(s) \in \mathcal{F}_{\text{NN}}, \quad \phi \sim P(\phi)
\end{equation}
where $\mathcal{F}_{\text{NN}}$ denotes the function class achievable by the network architecture, and $\phi$ represents the network weights.

The key distinction from prior work~\cite{hwangbo2019learning, jeong2019modelling} is that the proposed approach does not learn a single $\tau_\phi$ to match the true reality gap (which would require real-world data). Instead, policies are trained to be robust against \textit{distributions} of perturbation functions by randomizing $\phi$ per episode. The hypothesis is that $\mathcal{F}_{\text{NN}}$, being more expressive than $\mathcal{F}_{\text{DR}}$, provides broader coverage of potential reality gaps. The experimental results in Section~\ref{sec:Experiments} validate this hypothesis.

\subsection{Training Robust Policies via Joint Torque Space Perturbations}
To train a policy that generalizes across diverse perturbation functions, the weights $\phi$ are randomized at the beginning of each episode. The sampled $\tau_\phi(\cdot)$ takes the simulation state as input and outputs the perturbation at each time step. The perturbation is added during forward dynamics:
\begin{equation}
\begin{split}
(s_{t+1}, r_{t+1}) &= \textit{forward}(s_t, \tau_\pi(o_t) + \tau_\phi(s_{\text{simul},t})) \\
\tau_\pi(o_t) &\sim \pi_\theta(\cdot|o_t)
\end{split}
\end{equation}

In this work, $\tau_\pi(o_t)$ represents the nominal joint torques from by the policy $\pi_\theta$, and $\tau_\phi$ is modeled as a Multi-Layer Perceptron (MLP) with two hidden layers of 32 units with tanh activations. An additional $\tanh$ activation is applied at the output layer to softly clamp the output. A maximum perturbation coefficient $\sigma_{\text{lim}}$ is then multiplied with the final output to compute the joint torque space perturbation at each time step. For stable output distribution, the input vector $s_{\text{simul}}$ is normalized, while $\phi$ is randomly sampled from a Gaussian distribution with zero mean and standard deviation $\sigma_{\phi,k} = \sqrt{\frac{1.5}{n_{\text{in}} + n_{\text{out}}}}$, inspired by Xavier initialization~\cite{glorot2010xavier}. For practical implementation, $s_{\text{simul},t}$ is replaced with the privileged observation vector $o_{\text{priv},t}$, which contains both proprioceptive and ground-truth state information detailed in Section~\ref{subsubsec:observation_space}. 

The perturbation is computed as:
\begin{equation}
\tau_\phi(o_{\text{priv}}) = \sigma_{\text{lim}} \times \tanh(\text{MLP}(\hat{o}_{\text{priv}}))
\end{equation}

The observation vector is normalized by its running standard deviation estimate $\sigma_{o_{\text{priv}}}$, but not zero-centered. A zero-centered normalization would result in:
\begin{equation}
\hat{o}'_{\text{priv}} = \frac{o_{\text{priv}} - \mu_{o_{\text{priv}}}}{\sigma_{o_{\text{priv}}}}
\end{equation}
where $\mu_{o_{\text{priv}}}$ is the running mean estimate. Instead, the vector is normalized such that zero input remains zero after normalization:
\begin{equation}
\hat{o}_{\text{priv}} = o_{\text{priv}} / \sigma_{o_{\text{priv}}}
\end{equation}

Additionally, $\tau_\phi$ is designed with zero bias in all layers, ensuring that zero input always results in zero output. It is assumed that the reality gap is minimal when the robot is not in contact, not moving, not actuated, and in the default configuration. During training, only half of the parallel environments receive non-zero perturbations, while the remaining environments experience zero perturbation. This design allows the policy to learn from both perturbed and unperturbed dynamics simultaneously. Note that $\phi$ is not learned; rather, $\phi$ is randomly sampled at the beginning of each episode.

Since the robot has a floating base, the dynamics equation and the joint torque space perturbation injection method involve the virtual joints as well. Virtual joints are non-physical constructs introduced into robot models of floating-base robots that represent the motion of the robot base in the inertial frame. Therefore, injecting joint torque space perturbations includes perturbing the robot base. Although reality gaps in joint torque space include moment perturbations as well, in practice the robot base is perturbed only with force perturbations during training, under the assumption that the moment perturbations equivalent to reality gaps are negligible.

\subsection{Training Setup}
The proposed method is compared to the domain randomization (DR) baseline and the extended RFI (ERFI) baseline from~\cite{campanaro2024learning}. All policies are trained to produce natural and stable gait patterns while controlling the humanoid robot TOCABI to follow velocity commands $c_{\text{cmd}} = [v_{x,\text{cmd}}, v_{y,\text{cmd}}, \omega_{z,\text{cmd}}]$. TOCABI is a full-sized robot that stands 180\,cm tall and weighs approximately 100\,kg, using harmonic drives with a gear ratio of 100:1 as its actuators. Such a high gear ratio leads to lower torque transparency and higher uncertainty in actuator dynamics, resulting in more challenging sim-to-real transfer. Figure~\ref{fig:training_framework} illustrates the training framework used in this work.

The control policies are trained using the Proximal Policy Optimization (PPO) algorithm~\cite{schulman2017ppo}. In addition to the standard PPO losses, a gradient penalty loss $\mathcal{L}_{\text{grad}}$ from~\cite{chen2024lipschitz} is jointly optimized for smooth policy behavior. The total loss function is:
\begin{equation}
\mathcal{L}_{\text{total}} = \mathcal{L}_{\text{recon}} + \mathcal{L}_{\text{PPO}} + \lambda_{\text{grad}} \mathcal{L}_{\text{grad}}
\end{equation}
where $\mathcal{L}_{\text{recon}}$ is the reconstruction loss employed to ensure the latent representation captures the underlying dynamics of the environment. It is calculated as the mean squared error between the privileged observations $o_{\text{priv},t}$ and the observations reconstructed by the decoder from the latent vector.

The GRU-based encoder is trained end-to-end, receiving gradients from multiple loss terms. The reconstruction loss ensures the latent representation captures privileged state information, while the PPO surrogate loss allows the encoder to adapt its representations to improve policy performance. The gradient penalty is computed with full backpropagation through the encoder, encouraging the entire observation-to-action pipeline to exhibit smooth behavior with respect to input variations.

The actor network is modeled as an MLP with two hidden layers of 256 ELU units, and the critic network has hidden layers of [512, 512, 256] ELU units. The policy takes as input the current proprioceptive observation concatenated with a latent vector from a GRU-based encoder. The encoder consists of a single-layer GRU with 256 hidden units followed by an MLP with one hidden layer of 256 ELU units, compressing the observation history into a 24-dimensional latent representation. This latent is regularized through an auxiliary decoder (MLP with hidden layers [128, 128]) that reconstructs the privileged observations.

Samples for policy updates are collected using the GPU-accelerated simulator Isaac Gym~\cite{makoviychuk2021isaacgym}, simultaneously rolling out 4,096 agents in simulation and collecting 98,304 samples per policy update for a total of 10,000 policy updates. The maximum episode length is 20 seconds of simulation time, with early termination triggered when undesired contact occurs at robot links other than the foot links or when the base link height falls outside a threshold range.

Table~\ref{tab:domain_rand} shows the domain randomization parameters used across different training methods. Parameters are indicated by bullets: those used only in the DR baseline, those shared by DR and ERFI baselines, and delay randomization used by all methods including the proposed approach.

All methods incorporate delay randomization (up to 10\,ms) to simulate latencies between action computation and execution. The ERFI baseline additionally randomizes the motor constant ($\pm$20\%), which is multiplied by $\tau_\pi$ to form the final input joint torque in simulation. Since the real robot controls motor current rather than torque directly, this randomization is necessary and accounts for the torque bias injection from~\cite{campanaro2024learning}. The DR baseline further randomizes physical parameters including terrain friction, link mass and center of mass, actuator armature and damping, observation noise, and applies random pushes by setting the base link velocity in the transverse plane to random values.

Following~\cite{campanaro2024learning}, half of the parallelized environments are perturbed during training: with random force injection in the ERFI baseline and with state-dependent joint torque space perturbation in the proposed method. Perturbing only half of the environments is necessary for learning proper gait patterns. The maximum perturbation magnitudes are set to 50\,Nm for joint torques and 80\,N for base forces. 

\begin{table}[t!]
\caption{Domain randomization parameters across training methods. Bullets indicate which method uses each parameter. Subscripts: def=default, arm=armature, damp=damping, mot=motor, Prop.=Proposed Method.} 
\label{tab:domain_rand}
\centering
\setlength{\tabcolsep}{3pt}
\begin{tabular}{lccc|cc}
\toprule
\textbf{Parameter} & \textbf{DR} & \textbf{ERFI} & \textbf{Prop.} & \multicolumn{1}{c}{\textbf{Range}} & \textbf{Unit} \\
\midrule
Terrain friction & $\bullet$ & & & $[0.6, 1.4]\times f_{\text{def}}$ & \multicolumn{1}{c}{-} \\
Link mass & $\bullet$ & & & $[0.6, 1.4] \times m_{\text{def}}$ & kg \\
Link CoM & $\bullet$ & & & $[-0.03, 0.03] + x_{\text{CoM}}$ & m \\
Actuator armature & $\bullet$ & & & $[0.6, 1.4] \times I_{\text{arm}}$ & kg$\cdot$m$^2$ \\
Actuator damping & $\bullet$ & & & $[0.0, 2.9] + c_{\text{damp}}$ & Nm$\cdot$s/rad \\
Random push & $\bullet$ & & & $[0.0, 0.5]$ & m/s \\
Obs. bias/noise & $\bullet$ & & & \multicolumn{1}{c}{-} & \multicolumn{1}{c}{-} \\
\midrule
Motor constant & $\bullet$ & $\bullet$ & & $[0.8, 1.2] \times c_{\text{mot}}$ & \multicolumn{1}{c}{-} \\
\midrule
Delay & $\bullet$ & $\bullet$ & $\bullet$ & $[0.0, 10.0]$ & ms \\
\bottomrule
\end{tabular}
\end{table}

\section{Experimental Results}
\label{sec:Experiments}
This section validates the hypothesis that state-dependent joint torque perturbations enable superior robustness to complex reality gaps. Simulation experiments in MuJoCo and IsaacGym isolate specific sim-to-real gaps: unmodeled actuator dynamics, contact uncertainties, and widened range of randomized parameters. Real-world experiments on the TOCABI humanoid validate zero-shot transfer capabilities. Validation regarding the applicability of different control modalities (e.g., position) is detailed in the Appendix \ref{app:modality}.

\subsection{Simulation Experiments}

\subsubsection{Sim-to-Sim Transfer: Nominal Performance}
To verify that robustness enhancements do not compromise basic locomotion capabilities, command tracking performance is evaluated in a nominal environment with parameters derived from system identification and manufacturer specifications. Policies trained in IsaacGym are tested at maximum training command limits in MuJoCo to assess sim-to-sim transfer.

\begin{figure}[t!]
    \centering
    \includegraphics[width=\linewidth]{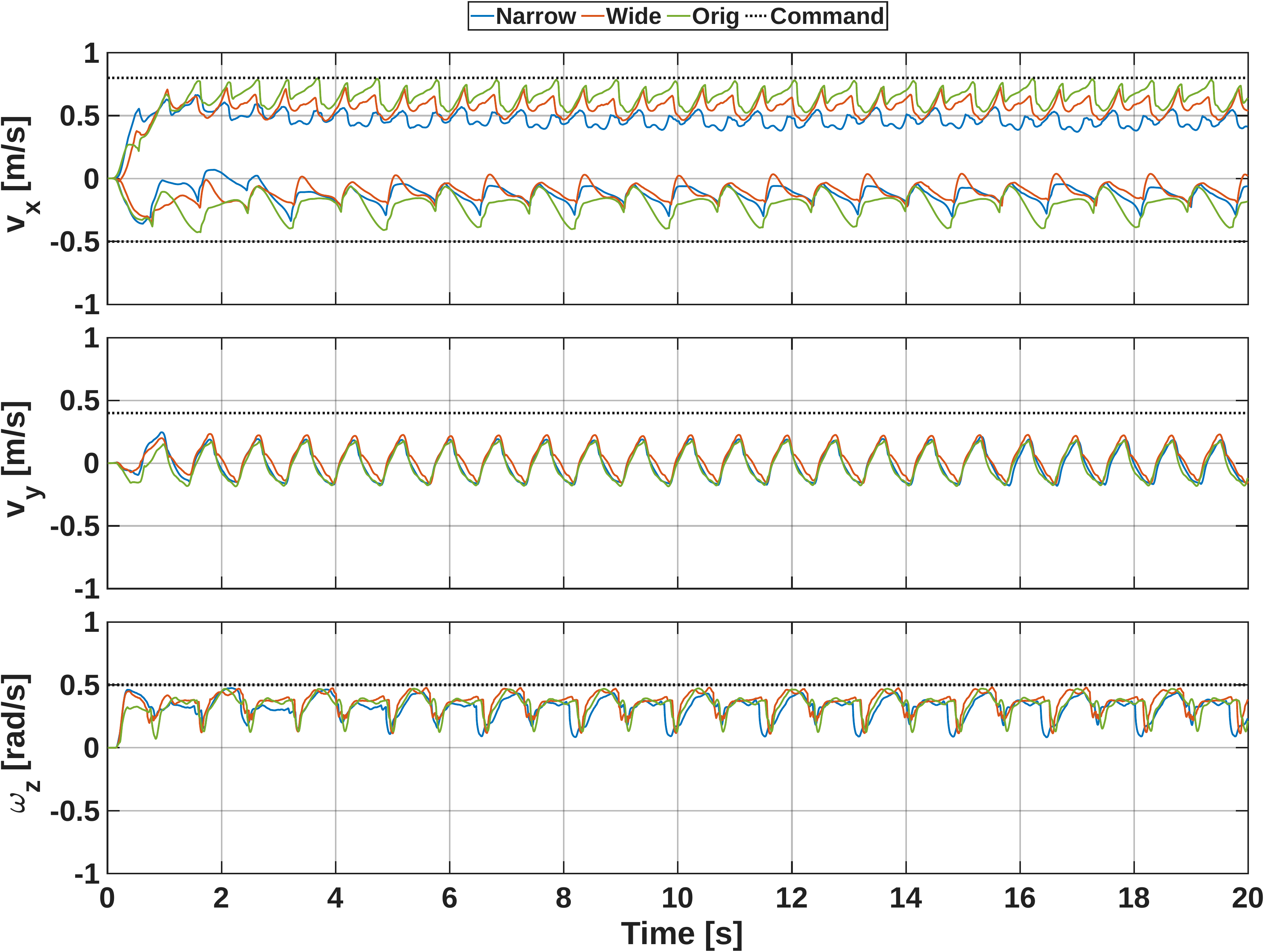}
    \caption{Command tracking performance in nominal MuJoCo environment at maximum training velocities (forward: $0.8\,\text{m/s}$, backward: $-0.5\,\text{m/s}$, lateral: $0.4\,\text{m/s}$, yaw: $0.5\,\text{rad/s}$).
    }
    \label{comparenominal}
\end{figure}

As shown in \autoref{comparenominal}, all methods achieve comparable tracking accuracy. RMSE values are [$0.2013$, $0.3298$, $0.3234$, $0.1925$], [$0.1449$, $0.2364$, $0.2676$, $0.2577$], and [$0.1337$, $0.1414$, $0.2280$, $0.2031$] for DR, ERFI baseline, and the proposed method respectively (forward, backward, lateral, yaw velocities in m/s, m/s, m/s, rad/s). This confirms that enhanced robustness does not sacrifice nominal performance.

\subsubsection{Robustness Against Unseen Actuator Dynamics}
This experiment tests whether neural perturbations can surrogate unmodeled physics that parametric DR cannot represent. A joint stiffness parameter ($250\,\text{Nm/rad}$) is introduced in the test environment, creating elastic spring elements at each joint with equilibrium at the initial configuration. This elasticity is never modeled during training. Standard DR with rigid-body parameter randomization cannot capture this compliance.

\begin{figure}[t!]
    \centering
    \includegraphics[width=\linewidth]{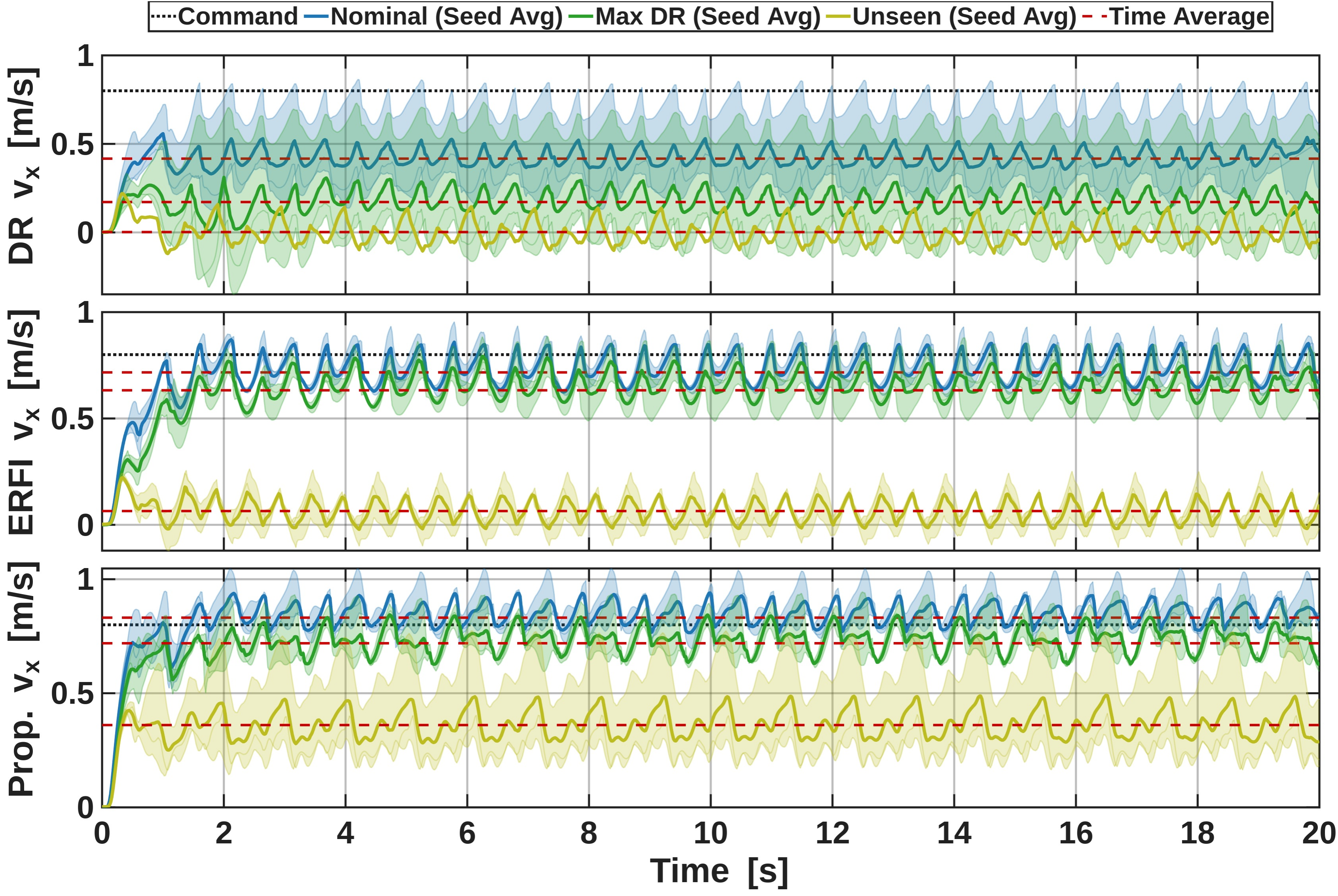}
    \caption{Command tracking performance ($0.8\,\text{m/s}$ forward) under nominal actuator, maximum DR, and unseen joint stiffness ($250\,\text{Nm/rad}$). Under stiffness, DR achieves $0.00\,\text{m/s}$ (catastrophic failure), ERFI $0.06\,\text{m/s}$ (91.9\% error), and proposed method $0.36\,\text{m/s}$ (54.9\% error), demonstrating superior robustness to unmodeled actuator dynamics.
    }
    \label{actcompare}
\end{figure}

Policies are evaluated under three conditions with $0.8\,\text{m/s}$ forward command: nominal parameters, maximum DR ranges for joint damping and friction loss, and added joint stiffness. As shown in \autoref{actcompare}, DR achieves $0.42\,\text{m/s}$ time-averaged velocity under nominal conditions, degrading to $0.17\,\text{m/s}$ under maximum DR ranges. When joint stiffness is introduced, DR catastrophically fails with two seeds collapsing completely and one achieving only $0.00\,\text{m/s}$ velocity.

The ERFI baseline achieves $0.72\,\text{m/s}$ under nominal conditions and $0.63\,\text{m/s}$ under maximum DR. However, when joint stiffness is introduced, performance collapses to $0.06\,\text{m/s}$ (91.9\% tracking error), demonstrating severe degradation despite maintaining gait generation. The proposed method achieves $0.83\,\text{m/s}$ under nominal conditions, $0.72\,\text{m/s}$ under maximum DR, and maintains $0.36\,\text{m/s}$ (54.9\% tracking error) despite joint stiffness, significantly outperforming both baselines under unmodeled elasticity. These results support the hypothesis that neural network-generated perturbations can represent dynamics mismatches that parametric randomization cannot capture.

\subsubsection{Robustness Against Unseen Contact Dynamics} \label{subsubsec:UnseenContactDynamics}
\begin{figure}[t!]
    \centering
    \includegraphics[width=\linewidth]{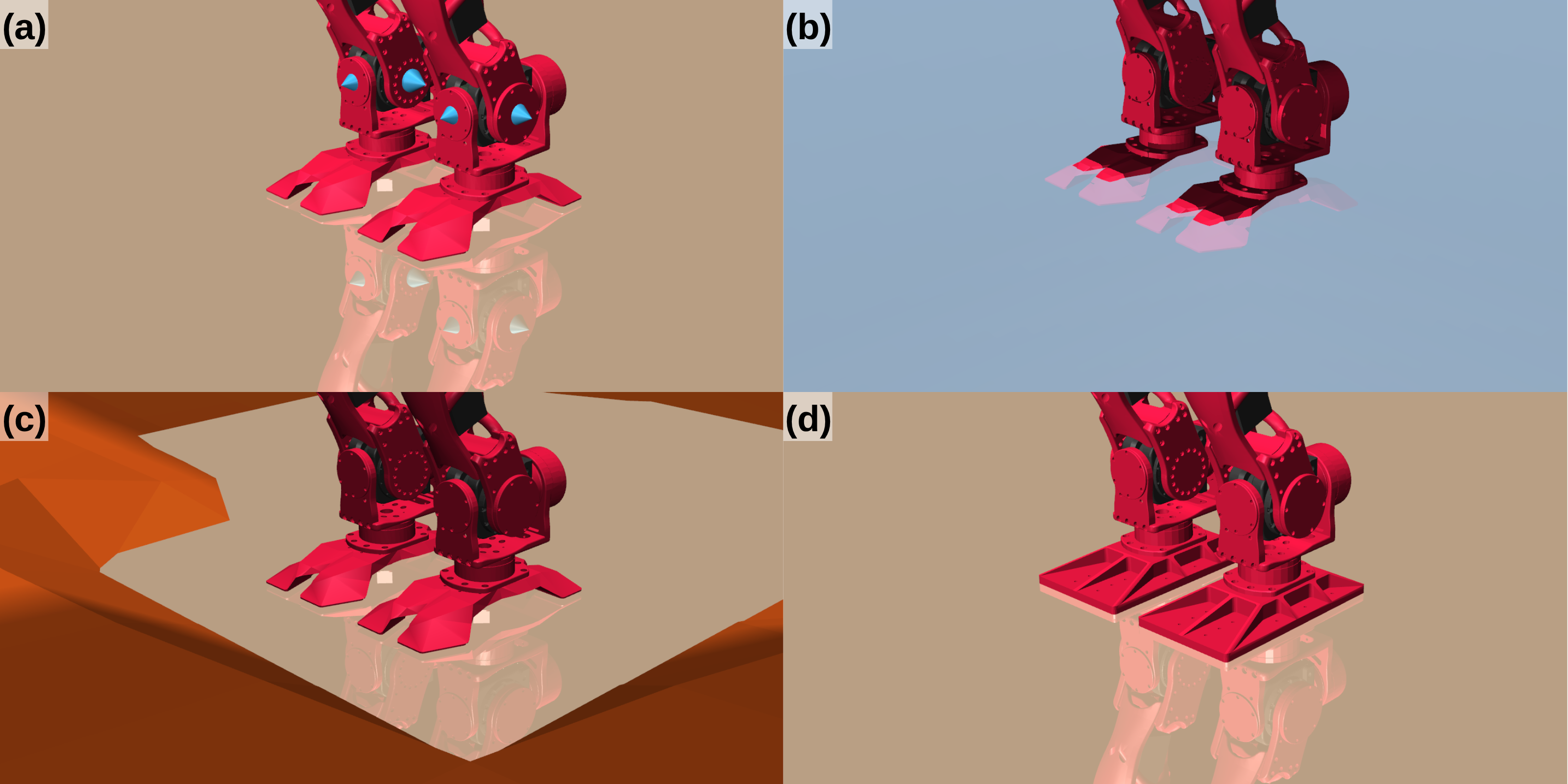}
    \caption{Test environment modifications. (Top left) Joint stiffness $250\,\text{Nm/rad}$ for actuator dynamics test. (Top right) Soft ground via solref time constant $0.1$. (Bottom left) Rough terrain from Gaussian-filtered noise. (Bottom right) Modified foot link geometry and mass. Contact dynamics tests combine top right, bottom left, and bottom right modifications.
    }
    \label{fig:solrefsoftenss}
\end{figure}

To evaluate robustness against contact model discrepancies, the test environment introduces soft ground contacts, unmodeled terrain variations, and foot link modifications. Ground contact stiffness is reduced by setting the solref time constant to $0.1$. Rough terrain is generated using random noise with Gaussian filtering. Foot link properties, such as shape and mass, are modified. All modifications are unmodeled during training. \autoref{fig:solrefsoftenss} shows how these modifications are emulated in MuJoCo.

\begin{figure}[t!]
    \centering
    \includegraphics[width=\linewidth]{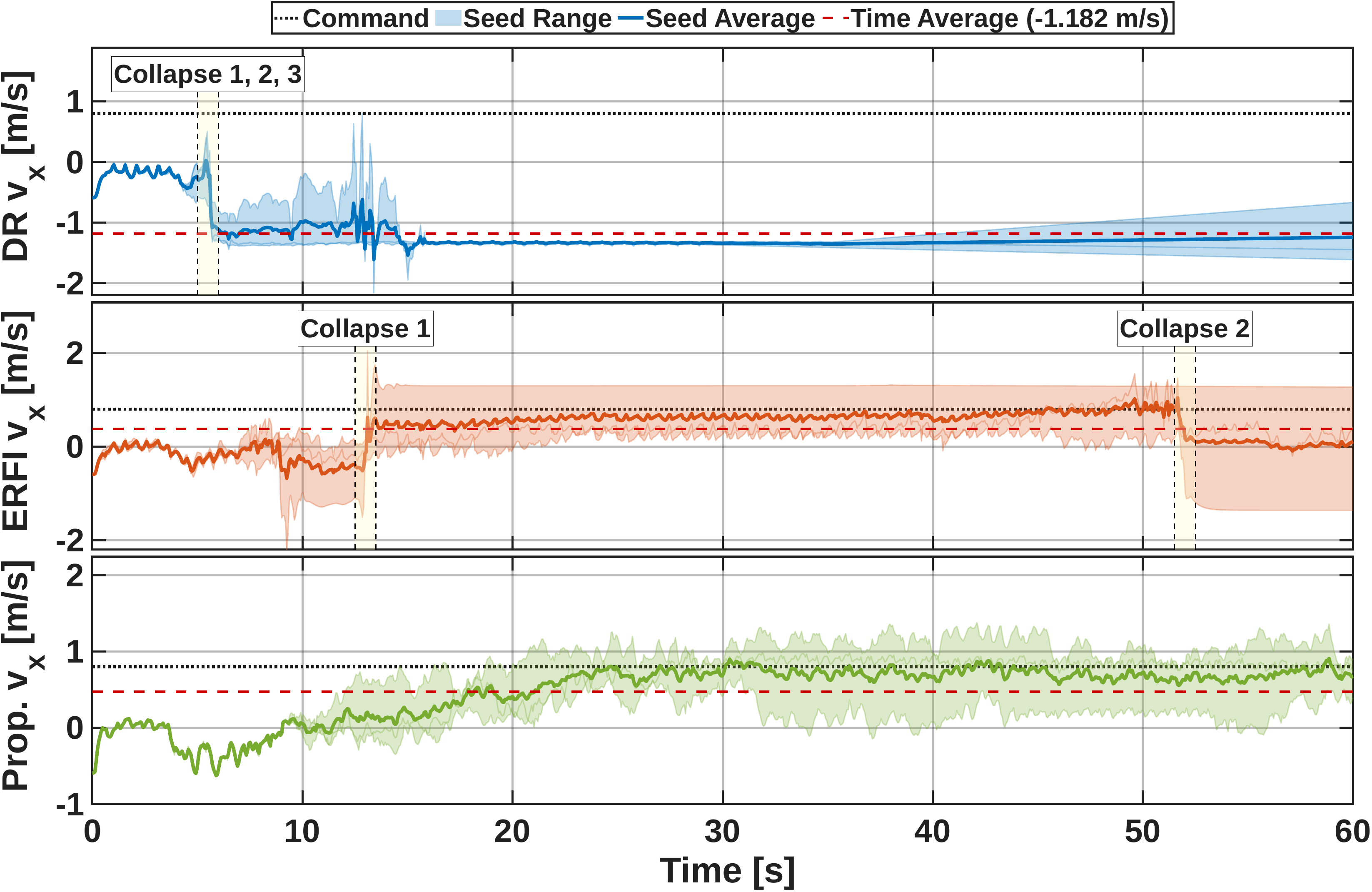}
    \caption{Command tracking performance ($0.8\,\text{m/s}$ forward) under unseen contact dynamics (soft ground, rough terrain, foot link changes). DR collapses before 6s ($-1.08\,\text{m/s}$). ERFI: two seeds collapse at 12s and 52s ($0.31\,\text{m/s}$ average). Proposed method: all seeds survive ($0.47\,\text{m/s}$, 40.9\% error).
    }
    \label{fig:envgap_compare}
\end{figure}

\autoref{fig:envgap_compare} shows command tracking performance over 60 seconds with $0.8\,\text{m/s}$ forward command across three seeds. All methods use identical terrain and starting positions. All three DR seeds collapse before 6 seconds. The ERFI baseline shows one seed collapsing at 12 seconds and another at 52 seconds. The proposed method maintains locomotion across all seeds. After 25 seconds, tracking performance improves significantly despite rough terrain sections.

\subsubsection{Widened Gap Test}
Policies are evaluated in IsaacGym under progressively challenging scenarios with 2048 parallel environments over 20-second episodes with a fixed $0.4\,\text{m/s}$ forward command. Critically, ERFI perturbations are disabled during evaluation to simulate deployment to reality, while sensor noise, bias, and external pushes remain enabled to stress test the policies.

\textbf{Scenario 1 (In-Distribution):} Policies are tested under the same randomization ranges used during training. This establishes baseline performance within the training distribution.

\textbf{Scenario 2 (OOD):} All randomization ranges are expanded $10\%$ beyond training limits, with ground friction variation introduced at $[0.7, 1.3]\times$ nominal, well outside training bounds (detailed ranges in Appendix \ref{app:dr_params}).

Table~\ref{tab:isaac_sim2real_gap} presents results across both scenarios. The DR baseline achieves $94.1\%$ success in-distribution but degrades to $89.5\%$ OOD, confirming hypothesized overfitting to training parameter bounds. Surprisingly, the ERFI baseline exhibits significant performance collapse, dropping from $82.3\%$ to $77.9\%$ success rate. This counterintuitive result, where the baseline trained with explicit perturbations performs worse than standard DR, suggests that inconsistent perturbation application during ERFI baseline training creates a detrimental train-test mismatch that undermines robustness.

In stark contrast, the proposed method achieves $99.7\%$ and $98.4\%$ success rates across both scenarios, demonstrating superior generalization with minimal performance degradation ($1.3\%$ drop vs. $4.6\%$ for DR and $4.4\%$ for ERFI baseline). The lateral drift metrics reveal comparable mean drift across methods ($\mu_y \approx -0.02$ to $0.03\,\text{m}$), with the proposed method maintaining trajectory consistency under OOD conditions.

\subsection{Real-World Experiments}
\label{subsec:Sim2Real}

\subsubsection{Zero-Shot Transfer Performance}
The trained policies are deployed directly on the TOCABI hardware without fine-tuning. Multiple training seeds of all methods (DR, ERFI, and the proposed method) successfully transfer to the real robot and track target velocity commands in a laboratory environment with uneven and slippery floors. The velocity commands consist of 16-second trajectories ramping from 0 to the target value and back to 0, with targets of 0.5\,m/s forward, 0.2\,m/s lateral, and 0.4\,rad/s yaw rotation.

\begin{figure}[t!]
    \centering
    \includegraphics[width=\linewidth]{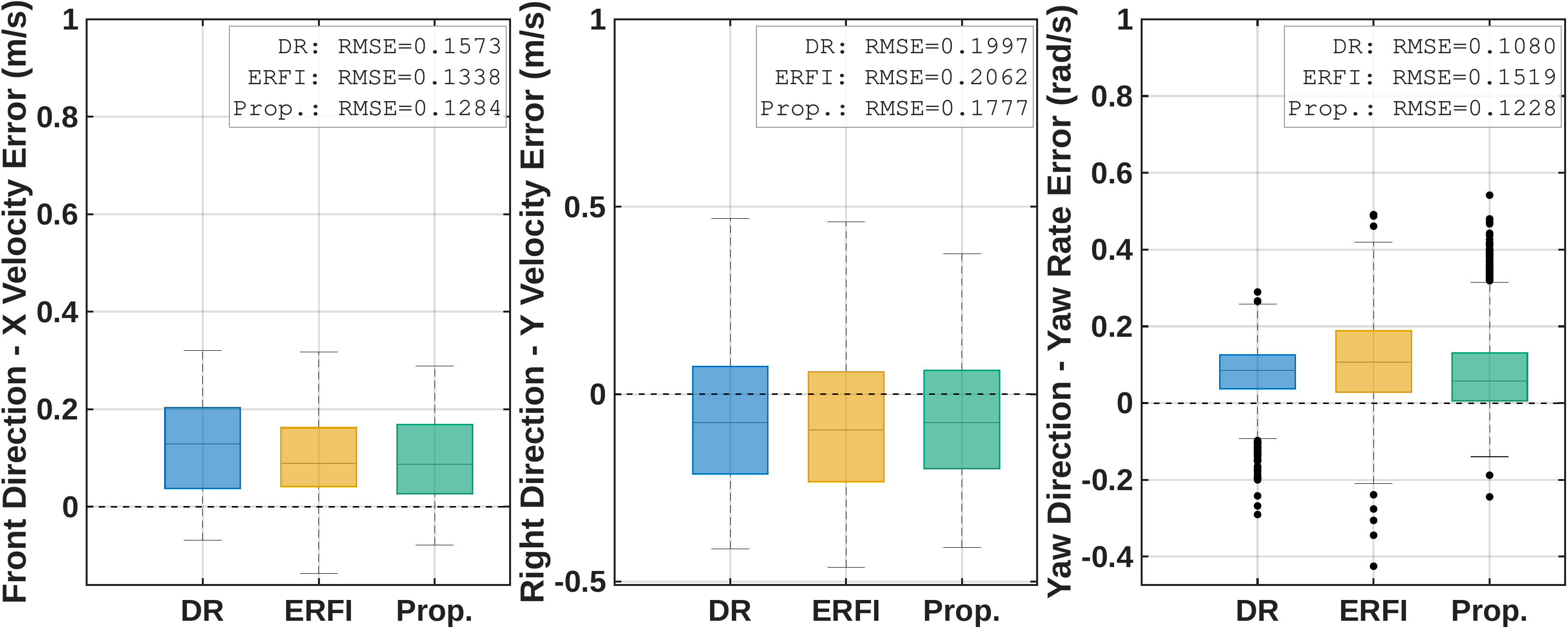}
    \caption{Box plots show median (center line), IQR (box), 1.5$\times$IQR whiskers, and outliers (dots).}
    \label{fig:error_distribution}
\end{figure}

\autoref{fig:error_distribution} shows command tracking error distributions for the three velocity components. All methods achieve comparable performance, with the proposed method showing slight advantages in forward (RMSE: 0.128\,m/s) and lateral (RMSE: 0.178\,m/s) tracking, while DR excels in yaw rate tracking (RMSE: 0.108\,rad/s).

\subsubsection{OOD Robustness}
\begin{figure}[t!]
    \centering
    \includegraphics[width=\linewidth]{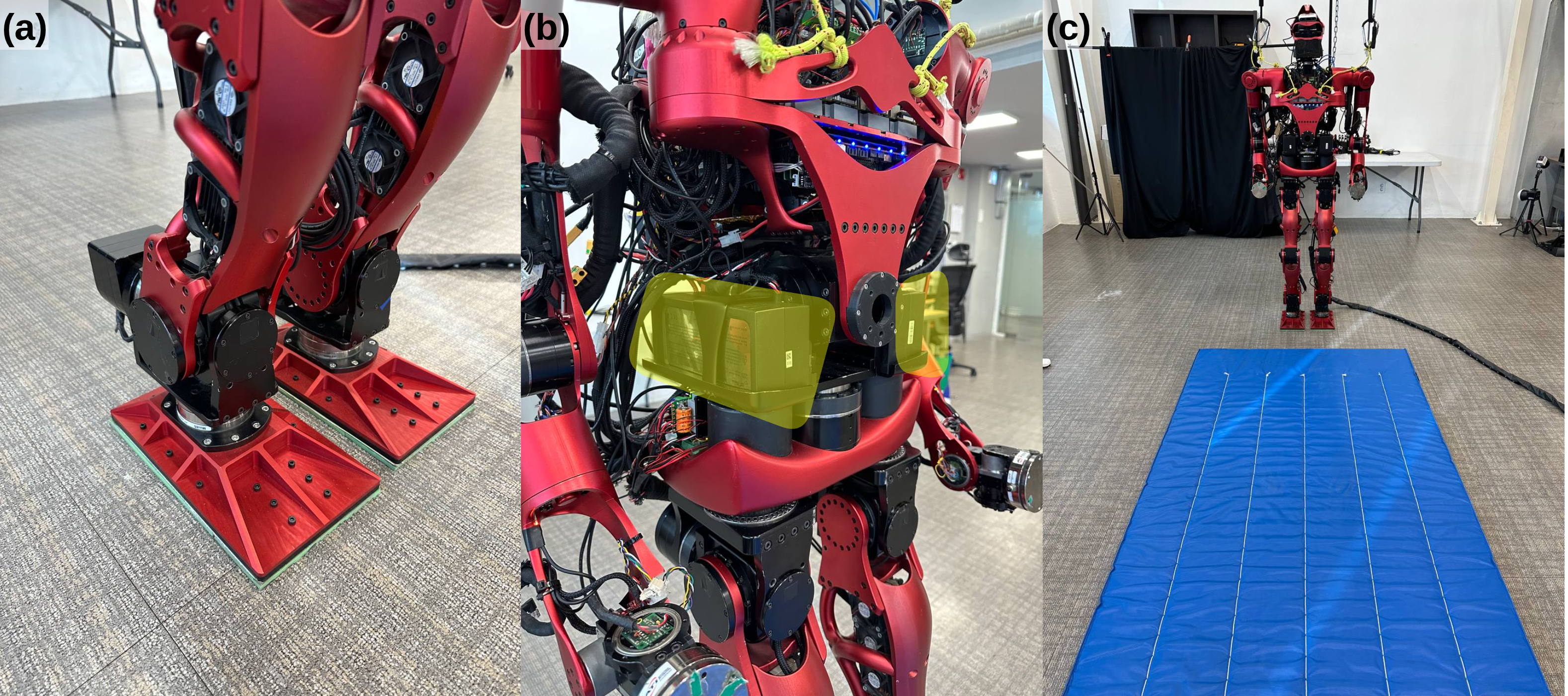}
    \caption{Hardware modifications for OOD testing. (a) Modified feet with different contact dynamics (+3\,kg per foot). (b) Additional batteries on base (+3\,kg each). (c) Compliant gym mat terrain (150$\times$200\,cm, 6\,cm thick).} 
    \label{fig:realrobot_snapshot}
\end{figure}

To evaluate robustness beyond standard conditions, intentional hardware modifications are introduced to create sim-to-real discrepancies, as illustrated in \autoref{fig:realrobot_snapshot}.

\textbf{Dynamic mismatch:} Two configurations alter the robot dynamics. \ref{fig:realrobot_snapshot}(a) shows the robot equipped with feet exhibiting different contact dynamics than those used during training, adding 3\,kg per foot. \ref{fig:realrobot_snapshot}(b) shows additional batteries mounted on the base, each adding approximately 3\,kg.

\textbf{Rough terrain locomotion:} Performance is evaluated on a 150\,cm $\times$ 200\,cm, 6\,cm thick gym mat, \ref{fig:realrobot_snapshot}(c), which presents compliant and deformable terrain more challenging than the laboratory floor. The robot tracks 0.4\,m/s forward velocity during this test.

\begin{figure}[t!]
    \centering
    \includegraphics[width=\linewidth]{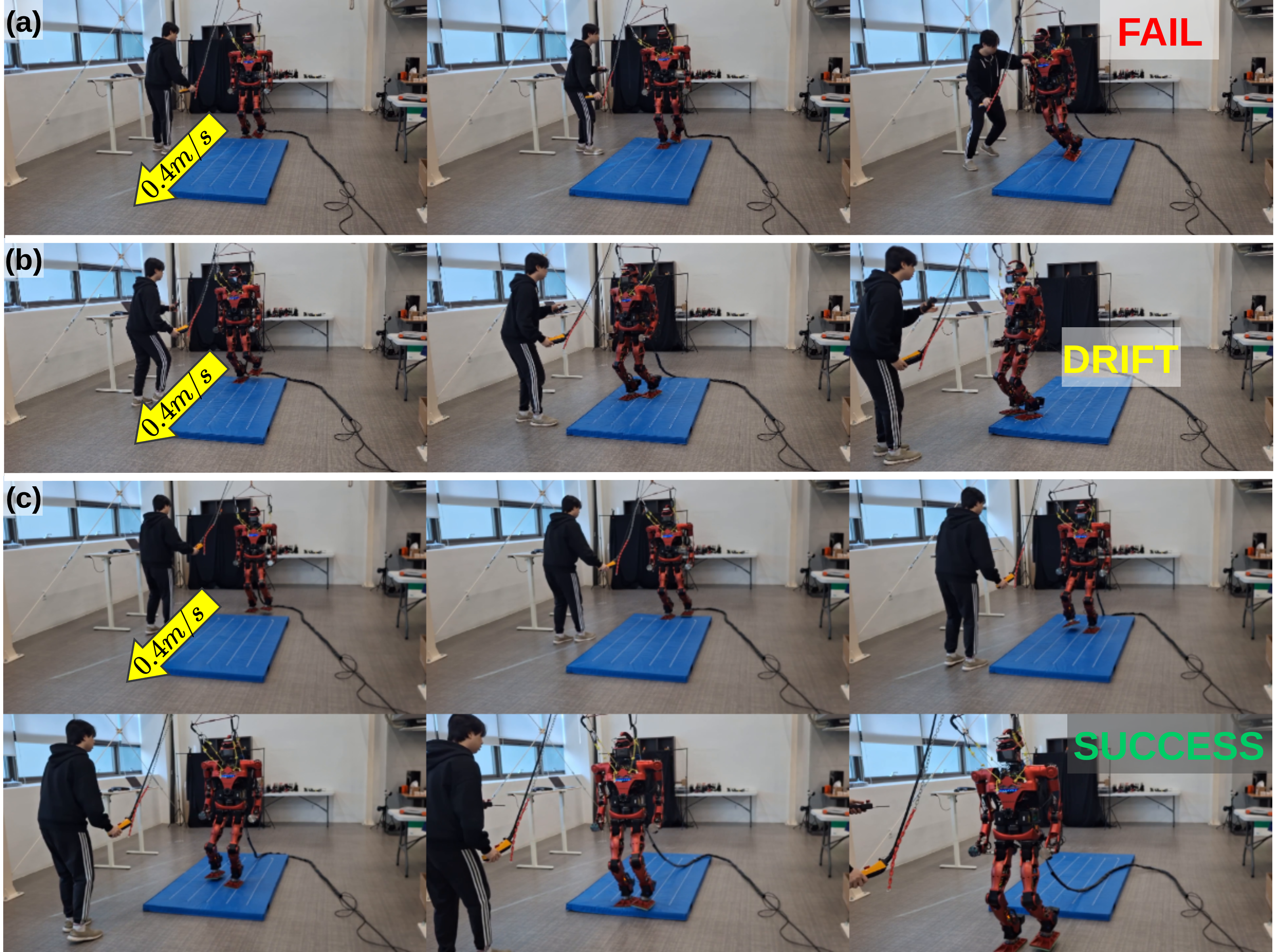}
    \caption{OOD real-world results. (a) DR: all seeds fail on rough terrain. (b) ERFI: partial success with some seeds failing or exhibiting severe yaw drift. (c) Proposed method: all seeds succeed, maintaining balance and minimizing drift.} 
    \label{fig:realrobot_ood}
\end{figure}

\autoref{fig:realrobot_ood} presents the results under these OOD conditions. The DR baseline, \ref{fig:realrobot_ood}(a) fails across all training seeds on the rough terrain, with the robot falling during forward locomotion. The ERFI baseline, \ref{fig:realrobot_ood}(b), shows partial success, with some seeds completing the task while others fail or exhibit severe yaw drift. The proposed method, \ref{fig:realrobot_ood}(c), demonstrates successful locomotion across all training seeds, maintaining balance and minimizing drift throughout the trajectory. Detailed tracking results for all seeds are provided in Appendix \ref{app:seeds}.

\begin{table}[t!]
\caption{Sim-to-real gap test in IsaacGym (2048 envs, 20s episodes, 0.4 m/s forward cmd).}
\label{tab:isaac_sim2real_gap}
\centering
\setlength{\tabcolsep}{5pt}
\begin{tabular}{lccc}
\toprule
\textbf{Method} & \textbf{Scenario} & \textbf{Success Rate} & \textbf{Drift $\mu_y \pm \sigma_y$ [m]} \\
\midrule
\multirow{2}{*}{DR} 
& \multicolumn{1}{l}{S1: Standard DR} & 94.1\% (1928/2048) & $-0.017 \pm 2.028$ \\
& \multicolumn{1}{l}{S2: Wider DR} & 89.5\% (1832/2048) & $0.008 \pm 2.009$ \\
\midrule
\multirow{2}{*}{ERFI} 
& \multicolumn{1}{l}{S1: Standard DR} & 82.3\% (1686/2048) & $0.033 \pm 1.860$ \\
& \multicolumn{1}{l}{S2: Wider DR} & 77.9\% (1596/2048) & $0.005 \pm 2.105$ \\
\midrule
\multirow{2}{*}{Prop.} 
& \multicolumn{1}{l}{S1: Standard DR} & \textbf{99.7\%} (2042/2048) & $0.009 \pm 2.206$ \\
& \multicolumn{1}{l}{S2: Wider DR} & 98.4\% (2015/2048) & $0.017 \pm 2.176$ \\
\bottomrule
\end{tabular}
\end{table}

\section{Conclusions} \label{conclusions}
This paper introduced a novel alternative to existing sim-to-real techniques by injecting state-dependent perturbations in joint torque space during control policy training. The key insight is that domain randomization can be equivalently viewed as perturbations in joint torque space, but with expressiveness limited by the simulator formulation. By replacing parametric randomization with neural network-generated perturbations, the proposed method can represent complex, state-dependent dynamics mismatches that fixed-parameter approaches cannot.% capture.

The proposed approach was validated through locomotion experiments on a full-sized humanoid robot, both in simulation and with real hardware. Simulation results indicate that the proposed method outperforms baseline techniques in addressing reality gaps arising from discrepancies that were not, or could not be, introduced during training, especially those regarding actuator and contact dynamics. In real robot experiments, the proposed method consistently showed enhanced robustness against the reality gap arising from current hardware and laboratory settings. These findings highlight the potential of functional perturbation approaches to generalize more effectively in cases where unmodeled dynamics or other complex factors play a critical role. Furthermore, these robustness gains did not come at the expense of task performance in unperturbed domains.

Overall, the proposed method offers promising pathways for bridging the reality gap in a wide array of robotic systems and tasks, extending beyond humanoid locomotion. It can further support the adoption of deep reinforcement learning algorithms in real-world environments, particularly for high-dimensional problems with steep sample complexity, by providing a robust framework for training policies that are inherently more adaptable to aspects of the dynamics that are not or cannot be modeled.

\appendices

% ====================================================================
% APPENDIX A: DEPLOYMENT & RAIBERT
% ====================================================================
\section{Deployment and Reference Generation}
\label{app:deployment_raibert}

\subsection{Deployment Architecture}
The deployment architecture leverages the DWL framework, which augments the standard PPO and value losses with a denoising objective consisting of ground-truth observation reconstruction and latent space regularization.

\begin{figure}[h]
    \centering
    \includegraphics[width=\linewidth]{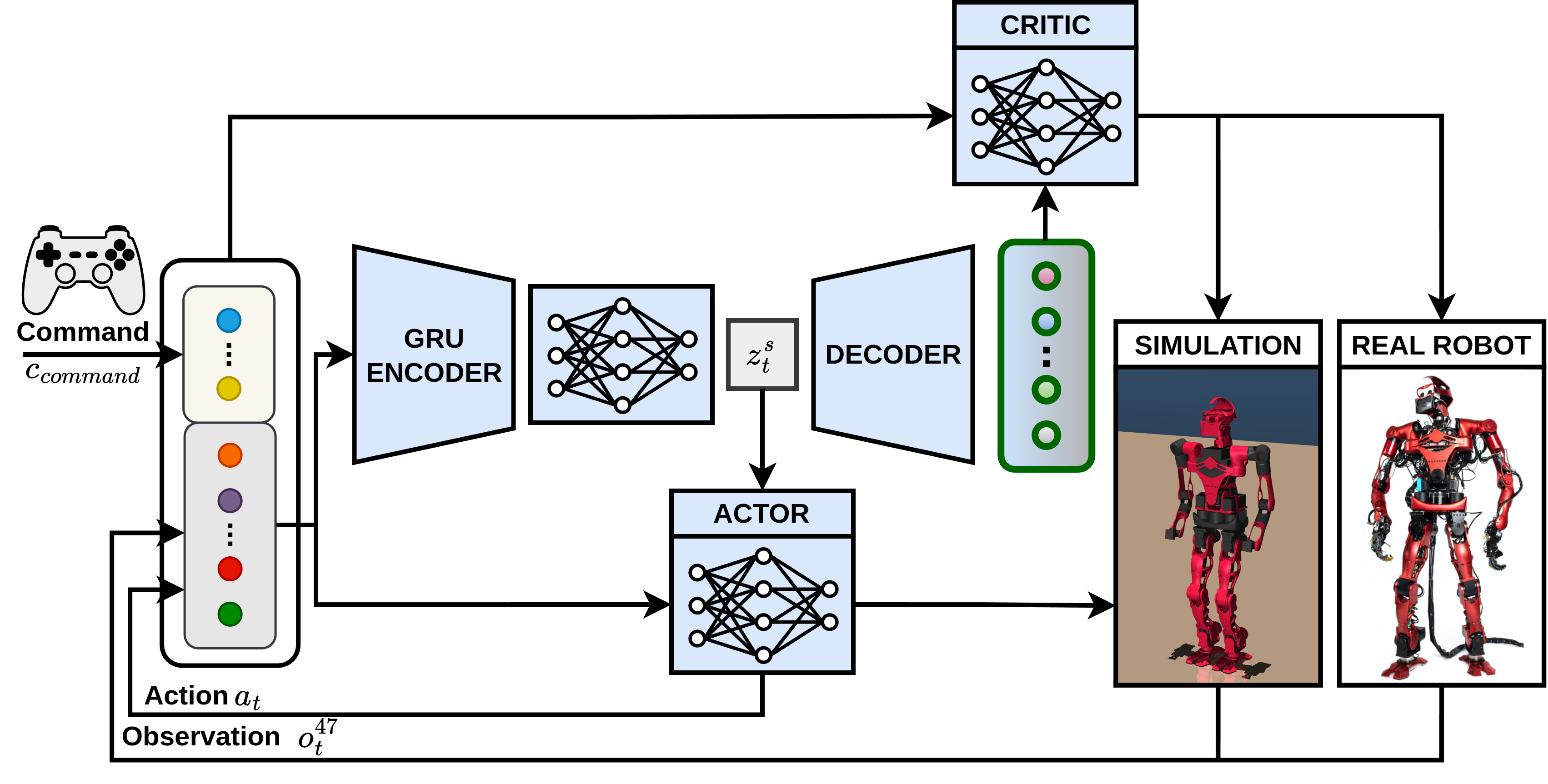}
    \caption{\textbf{Deployment Framework.} The controller receives velocity commands ($v_x$, $v_y$, $\omega_z$) from either joystick input or preset values.}
    \label{fig:test_framework}
\end{figure}

Figure~\ref{fig:test_framework} illustrates the inference pipeline for both simulation and physical hardware. Proprioceptive observations and user commands are processed by a GRU encoder to generate a latent representation. This latent vector is concatenated with the normalized observations and fed into the actor network to compute control actions. Simultaneously, the ground-truth observation, reconstructed via the decoder, is input into the critic alongside proprioceptive data to estimate the state value, serving as a runtime metric for the safety status of the robot.

\begin{figure}[h]
    \centering
    \includegraphics[width=\linewidth]{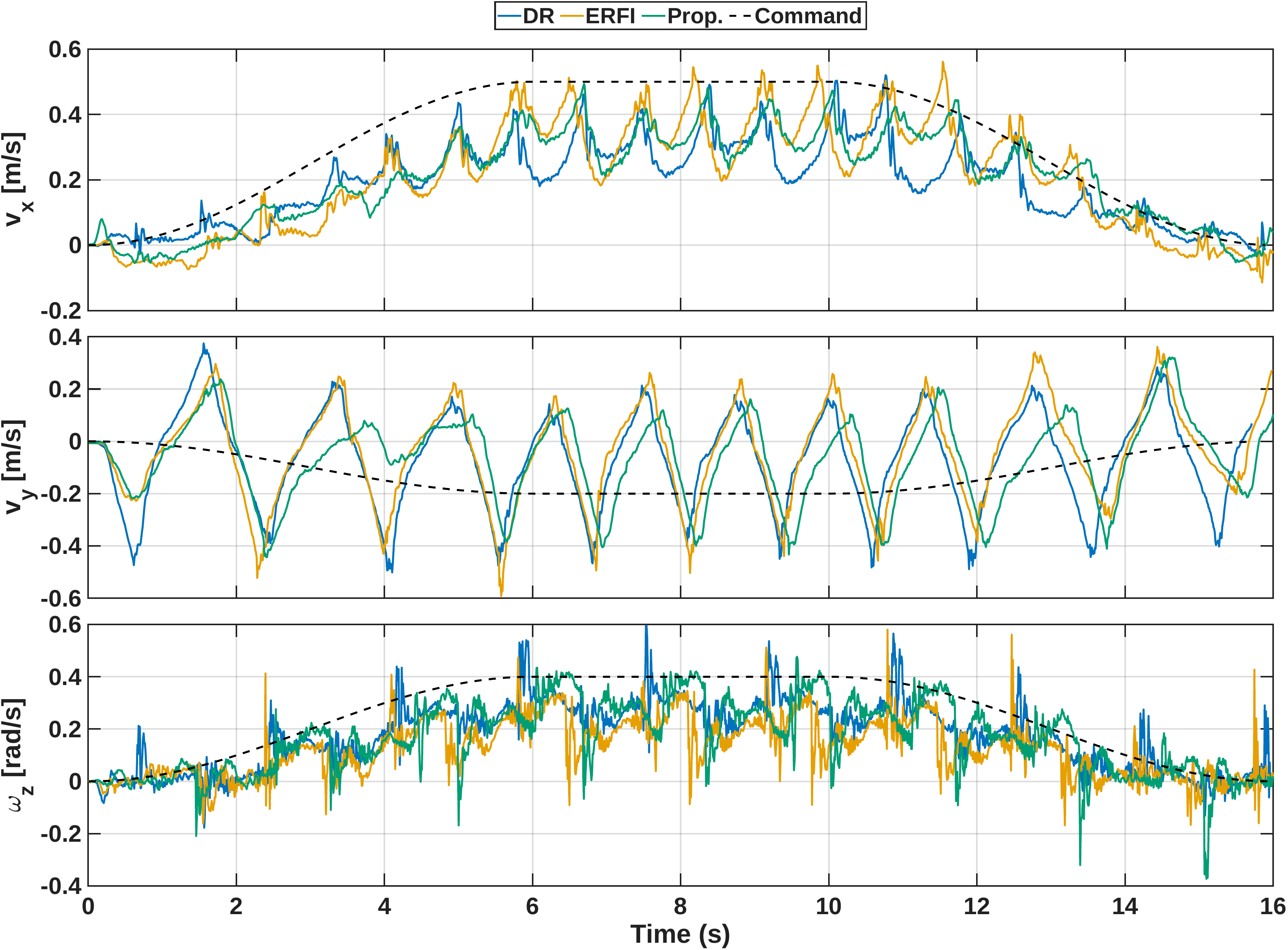}
    \caption{Velocity tracking performance on the real robot: forward, lateral, and yaw components. Linear velocities of the real robot are estimated by combining force-torque sensor measurements with forward kinematics.}
    \label{fig:real_vel_tracking_comparison}
\end{figure}

Figure~\ref{fig:real_vel_tracking_comparison} illustrates velocity tracking performance on the real robot for DR, ERFI, and the Proposed method across forward ($v_x$), lateral ($v_y$), and yaw ($\omega_z$) directions.

\begin{figure}[h]
    \centering
    \includegraphics[width=\linewidth]{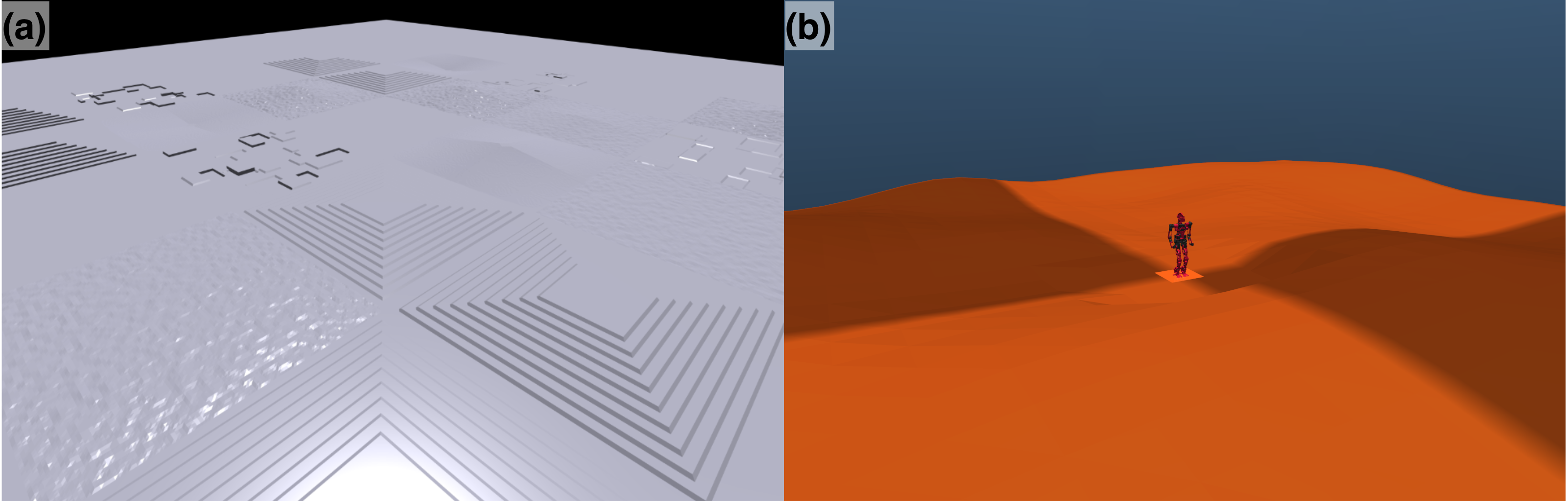}
    \caption{\textbf{Training vs. Testing Domains.} Visual comparison of the training (a) versus the testing environment (b).}
    \label{fig:testing_and_terrain_env}
\end{figure}

Figure~\ref{fig:testing_and_terrain_env} contrasts the nominal training environment with the OOD testing scenarios. The testing environments introduce unseen dynamics, including variable ground stiffness (modified \texttt{solref} parameters), rough terrain generation, and unmodeled foot contacts.

% ====================================================================
%  Raibert Heuristic Footstep Planning & Reference Trajectory Generation
%  - Drop this into your paper's methodology section.
%  - Requires: amsmath, amssymb, bm
% ====================================================================

\subsection{Footstep Planning and Swing Trajectory Generation}
\label{sec:footstep_planning}

To encourage natural gait characteristics without adversarial priors, a  customized Raibert-style heuristic~\cite{raibert1986legged} is employed to compute target footholds online and generate smooth reference swing-foot trajectories. The planner operates in a strictly alternating single-support gait where a binary phase indicator $\phi \in \{0,1\}$ encodes the stance leg ($\phi=0$: left; $\phi=1$: right).

\subsubsection{Step Period Adaptation}
Given the commanded body velocity $\bm{c_{command}}=(v_x^{\mathrm{cmd}},\,v_y^{\mathrm{cmd}},\,\omega_z^{\mathrm{cmd}})$, the step period $T_{\mathrm{step}}$ adapts dynamically to locomotion speed:
\begin{multline}
T_{\mathrm{step}}
= \mathrm{clip}\Bigl(
\min\Bigl\{
\frac{c_x}{\lvert v_x^{\mathrm{cmd}}\rvert+\epsilon},\;
\frac{c_y}{\lvert v_y^{\mathrm{cmd}}\rvert+\epsilon},\;
\frac{c_\psi}{\lvert\omega_z^{\mathrm{cmd}}\rvert+\epsilon}
\Bigr\}, \\
T_{\min},\;T_{\max}\Bigr)
\label{eq:step_period}
\end{multline}

where $c_x=c_y=c_\psi=0.4$ are threshold coefficients, $T_{\min}=0.4\,\text{s}$, $T_{\max}=0.8\,\text{s}$, and $\epsilon$ is a small stability constant. This formulation yields higher stepping frequencies at high velocities instead of increasing step length infinitely.

\subsubsection{Foothold Computation}
Target footholds are computed in the \emph{stance-foot frame} $\mathcal{S}$. The forward step length is proportional to velocity, clamped by $\Delta x_{\max}$:
\begin{equation}
\Delta x = \mathrm{clip}\bigl(T_{\mathrm{step}}\, v_x^{\mathrm{cmd}},\;
-\Delta x_{\max},\;\Delta x_{\max}\bigr).
\label{eq:raibert_x}
\end{equation}

Lateral placement maintains a minimum separation $w=0.21\,\text{m}$. For $v_y^{\mathrm{cmd}} \ge 0$:
\begin{equation}
\Delta y = \phi\bigl(0.5\,T_{\mathrm{step}}\,v_y^{\mathrm{cmd}} + w\bigr) - (1-\phi)\,w,
\end{equation}
and for $v_y^{\mathrm{cmd}} < 0$:
\begin{equation}
\Delta y = (1-\phi)\bigl(0.5\,T_{\mathrm{step}}\,v_y^{\mathrm{cmd}} - w\bigr) + \phi\,w.
\end{equation}
The target yaw offset is simply $\Delta\psi = T_{\mathrm{step}}\,\omega_z^{\mathrm{cmd}}$. 

\subsubsection{Swing Trajectory Optimization}
A continuous reference trajectory is generated over the normalized phase $s \in [0,1]$. Horizontal components ($x, y, \psi$) are interpolated via cubic splines satisfying boundary positions and velocities. The vertical component $z(s)$ is parameterized by a quartic polynomial to ensure toe clearance ($v_{\mathrm{lift}}$) and a specific apex height ($h_{\mathrm{apex}}$):
\begin{equation}
z(s) = \sum_{k=0}^{4} c_k\, s^k.
\end{equation}
The coefficients are solved offline to satisfy $z(0)=z(1)=0$, $z(0.5)=h_{\mathrm{apex}}$, $z'(0)=v_{\mathrm{lift}}$, and $z'(1)=0$.

\begin{figure}[h]
    \centering
    \includegraphics[width=\linewidth]{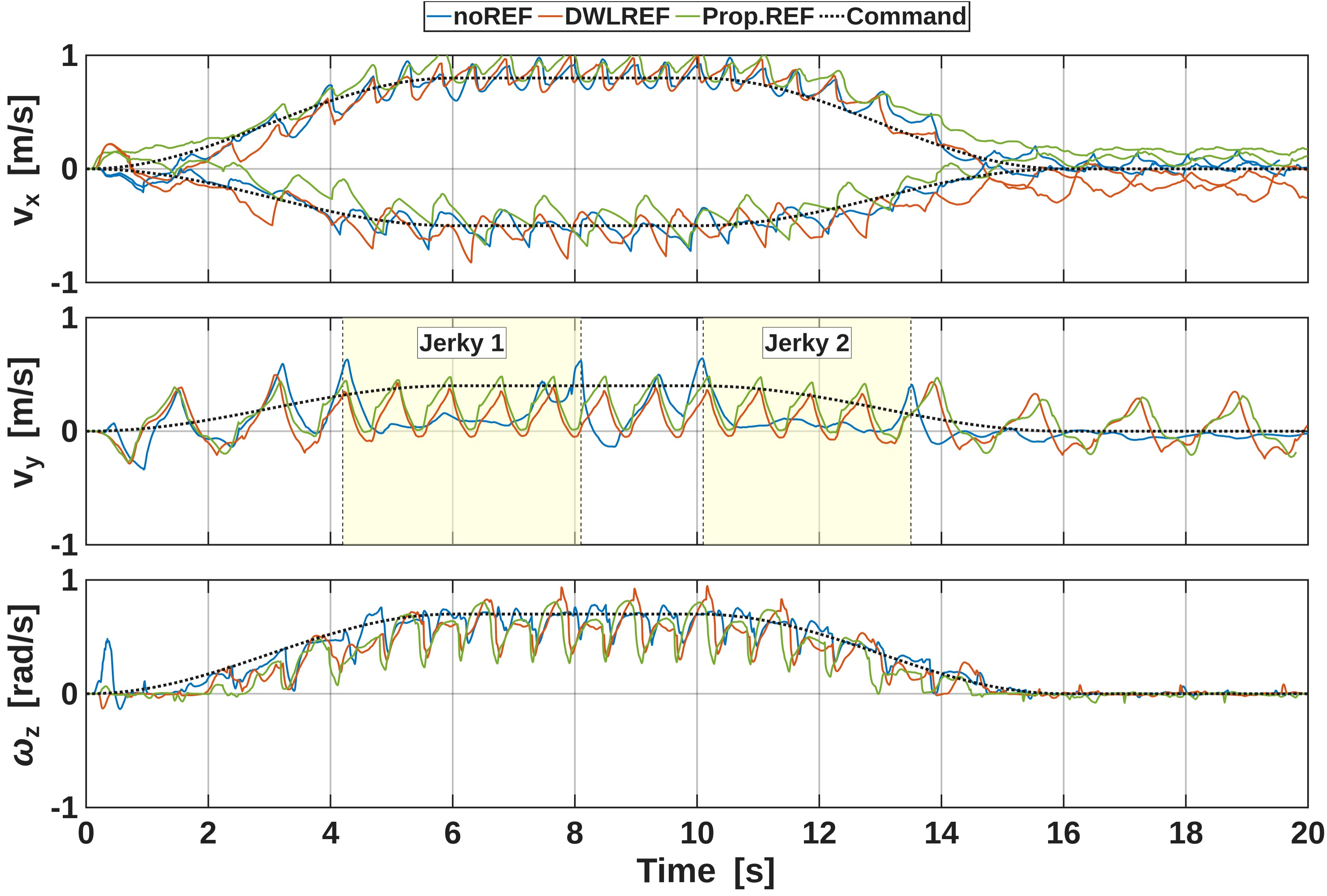}
    \caption{\textbf{Reference Trajectory Ablation.} Comparison of command tracking performance between noREF (no reference), DWLREF, and Prop.REF.}
    \label{fig:comparison_references_performance}
\end{figure}

As shown in Figure~\ref{fig:comparison_references_performance}, the noREF baseline (without reference trajectory) exhibits sudden jerks and imbalanced motion, particularly when tracking lateral velocity commands. Compared to the DWLREF approach, the Proposed method (Prop.REF) demonstrates the most natural and balanced locomotion across all velocity commands.

\label{app:raibert}

\begin{figure}[h]
    \centering
    \includegraphics[width=\linewidth]{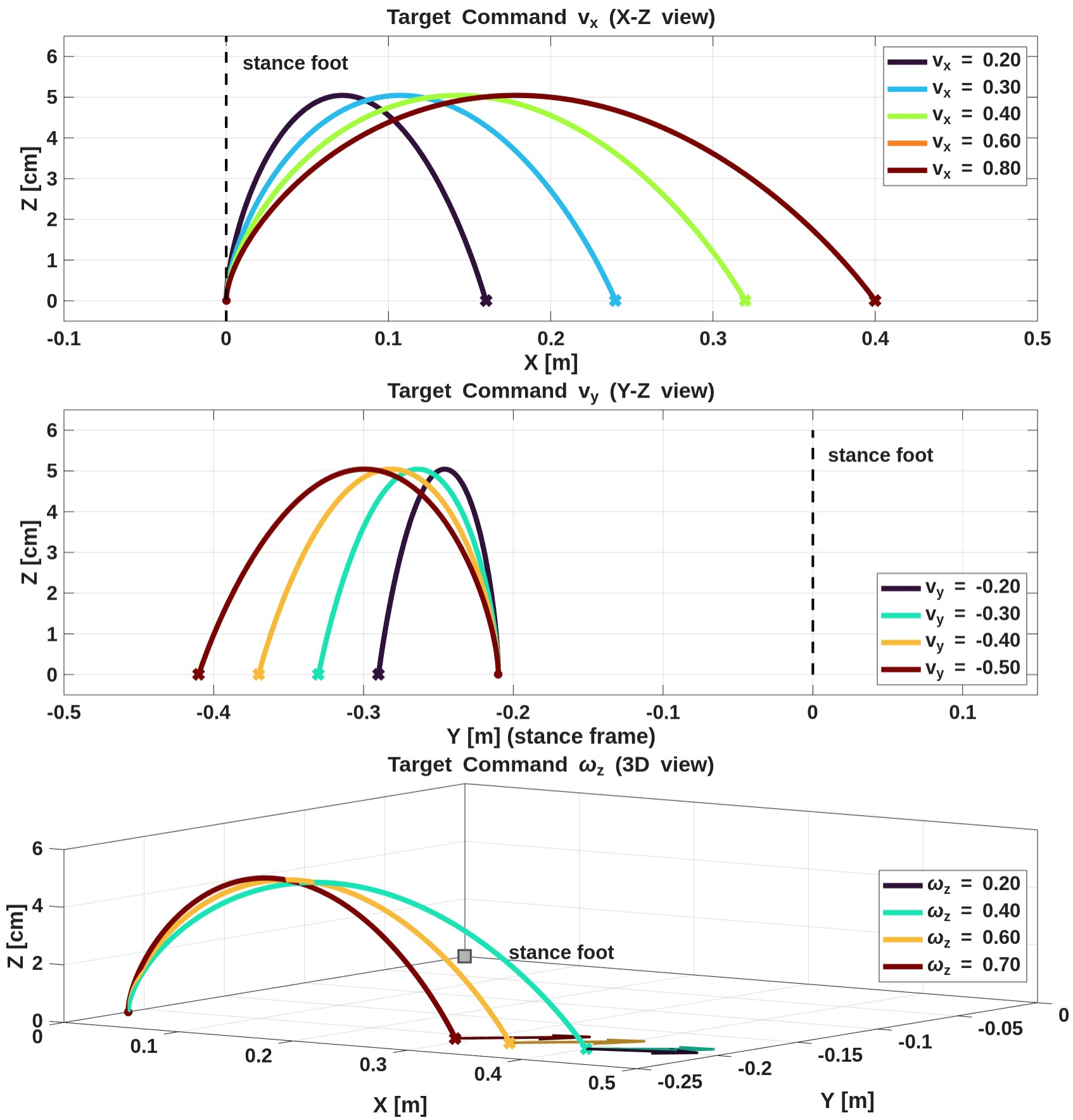}

    \caption{\textbf{Swing foot trajectories under velocity commands}. To ensure stable velocity tracking, the step period $T$ decreases for high-speed commands that would exceed kinematic limits, but is bounded by a minimum threshold of $T_{\min}=0.4$\,s (e.g., $T=0.67$\,s at $v_x=0.6$\,m/s, $T=0.5$\,s at $v_x=0.8$\,m/s). This dual constraint maintains both kinematic feasibility and gait stability.}
    \label{fig:app_raibert_traj}
\end{figure}

% ====================================================================
% APPENDIX B: REWARDS
% ====================================================================

\section{Rewards and Weights}
\label{app:rewards}
The reward function $r_{\text{total}} = \sum_{i} w_i r_i$ combines tracking objectives, motion quality heuristics, and regularization terms. Detailed reward components used are summarized in Table~\ref{tab:rewards}

\begin{table}[h]
\centering
\caption{Reward Function Components for Robust Locomotion}
\label{tab:rewards}
\setlength{\tabcolsep}{3pt}
\renewcommand{\arraystretch}{1.3}
\footnotesize
\begin{tabular}{@{}lccr@{}}
\toprule
\textbf{Component} & \textbf{Formulation} & \textbf{$\omega$} & \textbf{$\sigma$} \\
\midrule
\multicolumn{4}{@{}l}{\textit{\textbf{Command Tracking}}} \\
Linear Vel. X & $\exp(-(\mathbf{v}_{\text{cmd}}^{x} - \mathbf{v}_{\text{base}}^{x})^2/\sigma_{x})$ & 1.2$^{\dagger}$ & 0.15 \\
Linear Vel. Y & $\exp(-(\mathbf{v}_{\text{cmd}}^{y} - \mathbf{v}_{\text{base}}^{y})^2/\sigma_{y})$ & 1.2$^{\dagger}$ & 0.15 \\
Angular Vel. Z & $\exp(-(\omega_{\text{cmd}}^{z} - \omega_{\text{base}}^{z})^2/\sigma_{\omega})$ & 2.2 & 0.10 \\
Yaw Drift & See Eq.~\eqref{eq:yaw_drift} & 0.4 & 0.09 \\
\addlinespace[3pt]
\hline
\multicolumn{4}{@{}l}{\textit{\textbf{Motion Quality}}} \\
Base Height & $\exp(-(h_{\text{base}} - 0.92\,\text{m})^2/\sigma_{h})$ & 2.0 & 0.15 \\
Orientation & See Eq.~\eqref{eq:orientation} & 1.6 & 0.15$^{\ddagger}$ \\
Roll Stability & $\exp(-(r/0.1)^2 - (\dot{r}/0.2)^2)$ & 1.4 & 0.10$^*$ \\
Smooth Motion & $\exp(-\|\ddot{\boldsymbol{\omega}}_{\text{base}}^{xy}\|^2/(5.0)^2)$ & 0.6 & 5.0 \\
Swing Foot Pos. & $\exp(-\|\mathbf{p}_{\text{sw}}^{\text{ref}} - \mathbf{p}_{\text{sw}}\|^2/\sigma_{p})$ & 0.45 & 0.2/0.05$^{\S}$ \\
Swing Foot Ori. & $\exp(-(\psi_{\text{sw}}^{\text{ref}} - \psi_{\text{sw}})^2/\sigma_{\psi})$ & 0.45 & 0.10 \\
Stance Foot Pos. & $\exp(-\|\mathbf{p}_{\text{st}}^{\text{ref}} - \mathbf{p}_{\text{st}}\|^2/\sigma_{p})$ & 0.25 & 0.2/0.1$^{\S}$ \\
Stance Foot Ori. & $\exp(-(\psi_{\text{st}}^{\text{ref}} - \psi_{\text{st}})^2/\sigma_{\psi})$ & 0.25 & 0.15 \\
Contact Schedule & Phase-aware contact reward & 0.9 & -- \\
Force Symmetry & $\exp(-|\|F_{\text{left}}\| - \|F_{\text{right}}\||/\sigma_{\text{sym}})$ & 0.7 & 0.25 \\
Joint Deviation & $\exp(-\|\mathbf{q} - \mathbf{q}_{\text{default}}\|^2/\sigma_q)$ & 0.18 & 2.0 \\
\addlinespace[3pt]
\hline
\multicolumn{4}{@{}l}{\textit{\textbf{Regularization}}} \\
Action Rate & $-\sum_{i}(\mathbf{a}_{t,i} - \mathbf{a}_{t-1,i})^2/\sigma_i^2 \cdot s(\mathbf{c})$ & $-0.001$ & $^{\|}$ \\
Energy & $-\sum_{i} \tau_i \dot{q}_i$ & $4\times10^{-4}$ & -- \\
Joint Limits & $-\sum_{i} \max(0, |\mathbf{q}_i| - \mathbf{q}_{\text{limit},i})$ & $-0.4$ & -- \\
Contact Power & $-\sum_{i} |F_i \cdot \mathbf{v}_i|$ & $-0.015$ & -- \\
Impact Force & See Eq.~\eqref{eq:impact_force} & $-0.5$ & -- \\
Landing Velocity & See Eq.~\eqref{eq:landing_velocity} & $-2.0$ & -- \\
\bottomrule
\end{tabular}
\vspace{0.05cm}
\begin{flushleft}
\scriptsize
$^\dagger$Combined weight 2.4: 50\% X-tracking + 50\% Y-tracking. \\
$^\ddagger$Orientation penalizes roll ($\sigma=0.15$), pitch ($\sigma=0.2$), and yaw coordination ($\sigma=0.25$). \\
$^*$Roll stability: angle tolerance $\sigma_r=0.1$ rad, rate tolerance $\sigma_{\dot{r}}=0.2$ rad/s. \\
$^{\S}$Horizontal (XY-plane), vertical (Z-axis) tolerances. \\
$^{\|}$Action rate: joint-dependent $\sigma$ (hip yaw: 0.9, knee: 0.55, ankle: 0.45) scaled by command magnitude. \\
Dash (--) indicates no $\sigma$ parameter. $s(\mathbf{c})$ is command-dependent scaling. \\
Heuristic references are computed online using Raibert-style foothold planning.
\end{flushleft}
\end{table}

\begin{equation}
\label{eq:yaw_drift}
r_{\text{yaw}} = \begin{cases}
-(1-\exp(-\omega^2_{\text{base},z}/\sigma_\text{yaw}^2)) & \text{if } |\omega_{\text{cmd}}^z| < 0.1 \text{ rad/s} \\
0 & \text{otherwise}
\end{cases}
\end{equation}

\begin{equation}
\label{eq:orientation}
r_{\text{ori}} = \exp\left(-\frac{r^2}{\sigma_r^2} - \frac{p^2}{\sigma_p^2} - \frac{\Delta\psi^2}{\sigma_{\psi}^2}\right)
\end{equation}

\begin{equation}
\label{eq:impact_force}
r_{\text{impact}} = \begin{cases}
-\max(0, \|F_{\text{foot}}\| - 1100\,\text{N})^2 & \text{if landing detected} \\
0 & \text{otherwise}
\end{cases}
\end{equation}

\begin{equation}
\label{eq:landing_velocity}
r_{\text{landing}} = \begin{cases}
-\max(0, |\dot{z}_{\text{foot}}| - 1.5\,\text{m/s}) & \text{if } z_{\text{foot}} < 0.1\,\text{m} \\
0 & \text{otherwise}
\end{cases}
\end{equation}

% ====================================================================
% APPENDIX C: MODALITY
% ====================================================================
\section{Control Modality Independence}
\label{app:modality}

The proposed sim-to-real strategy injects perturbations into the joint torque space subsequent to the input torque computation stage $\tau_{\text{input},t}$. Consequently, the method is theoretically independent of the chosen control modality (e.g., position vs. torque control).

\begin{figure}[h]
    \centering
    \includegraphics[width=\linewidth]{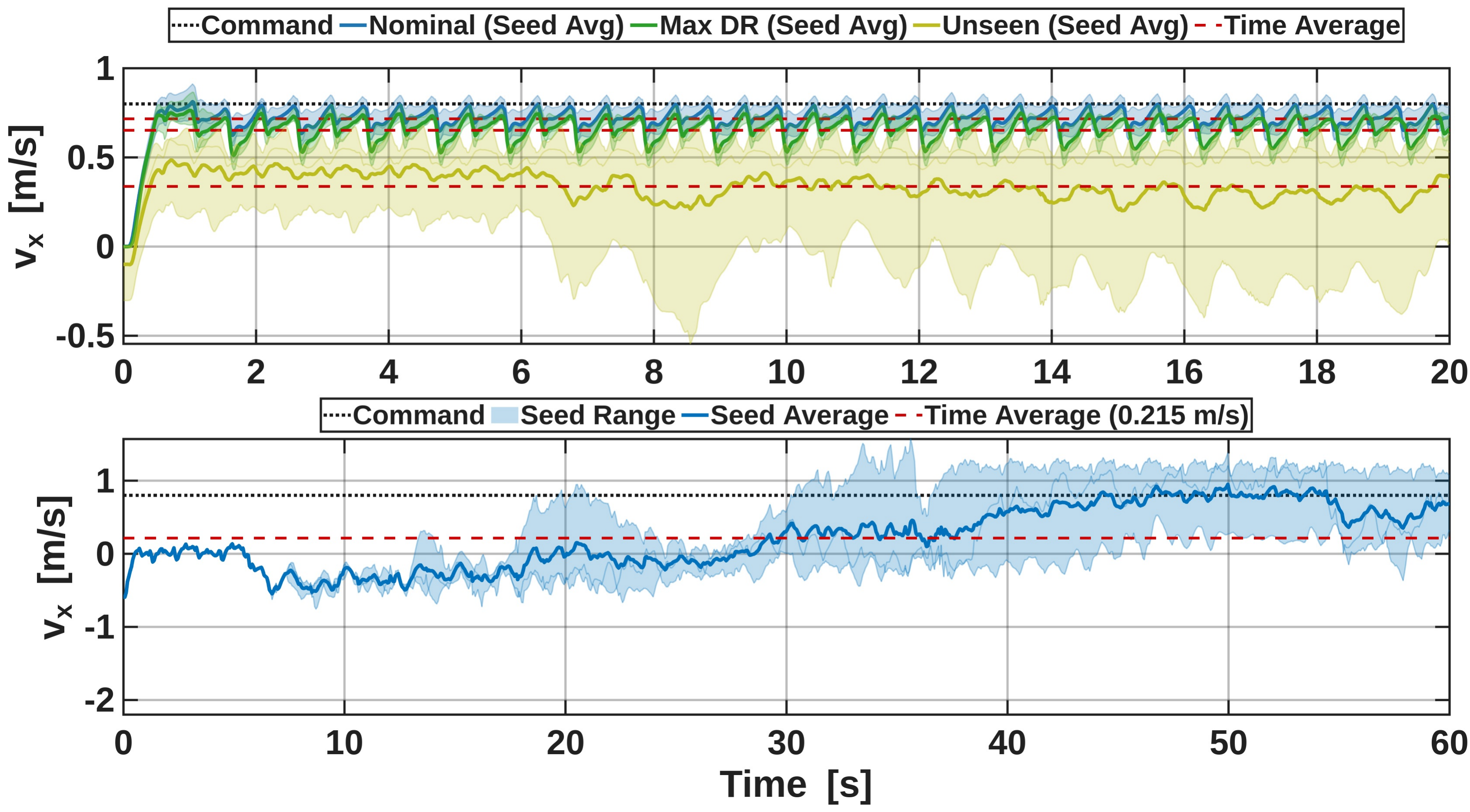}
    \caption{\textbf{Modality Robustness.} Evaluation of position-based control policies under unseen dynamics.}
    \label{fig:appendix_modality_robustness_all_merged}
\end{figure}

Figure~\ref{fig:appendix_modality_robustness_all_merged} presents experimental validation using a position-based controller across multiple seeds. The results confirm that the perturbation-based training effectively bridges the reality gap for position-controlled policies under unseen actuator and contact dynamics. However, it is acknowledged, as discussed in~\cite{kim2023torque}, that the performance of the position controller remains inherently sensitive to the tuning of PD gains.

% ====================================================================
% APPENDIX D: Domain Randomization
% ====================================================================

\section{Domain Randomization}
\label{app:dr_params}
Randomization ranges were modified from \cite{kim2024bridging} to optimize the trade-off between robustness and nominal tracking accuracy. While expanding randomization ranges generally improves OOD robustness, excessively wide ranges can degrade performance in nominal conditions due to the conservative behaviors learned by the policy.

\begin{figure}[h]
    \centering
    \includegraphics[width=\linewidth]{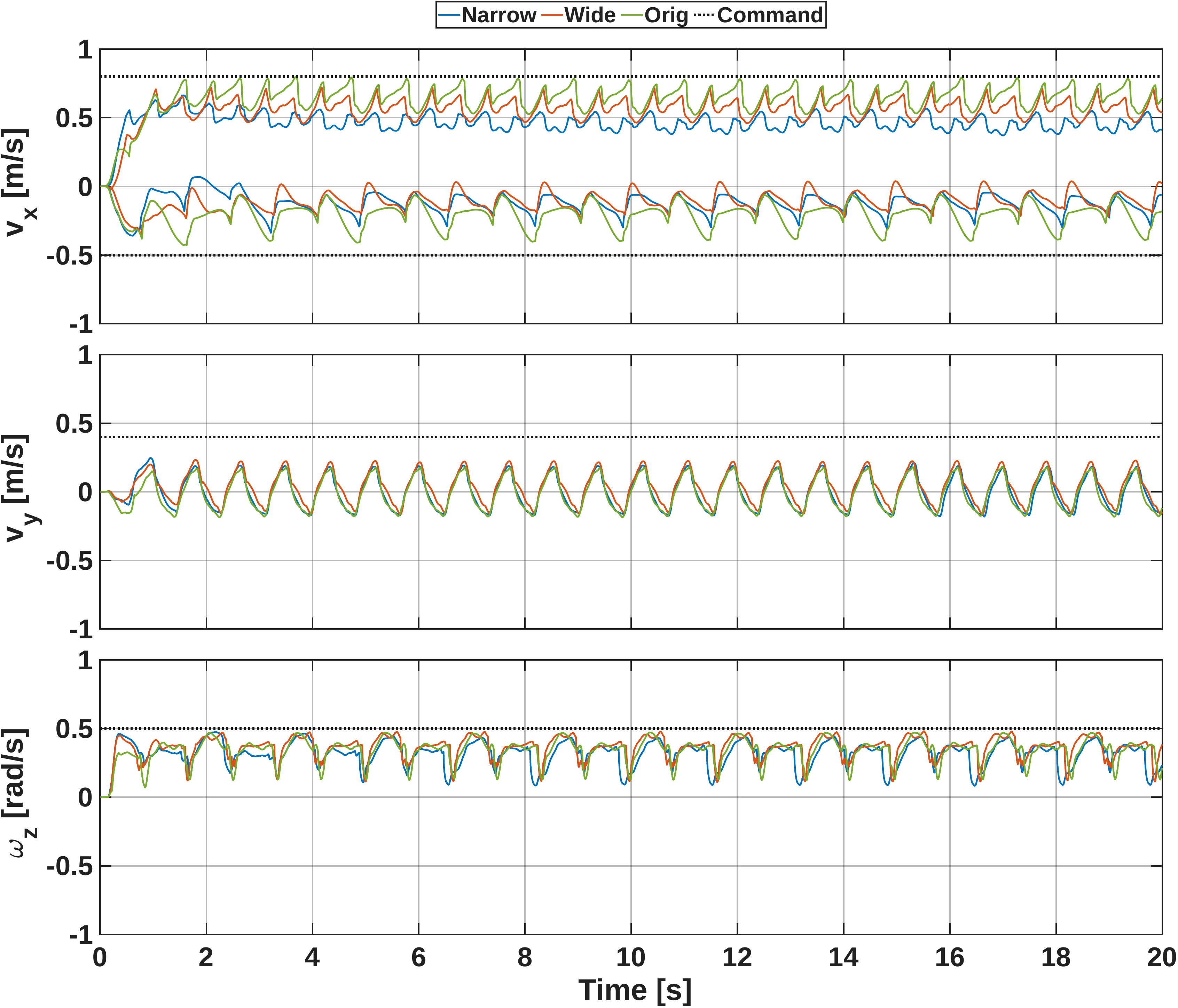}
    \caption{\textbf{Randomization Range Analysis.} Comparison of tracking performance across {Narrow}, {Wide}, and {Reference} randomization ranges.}
    \label{fig:randomization}
\end{figure}

As detailed in Table~\ref{tab:domain_randomization} and Table~I in the main paper, three randomization strategies were evaluated: \textbf{Narrow}, \textbf{Wide} and \textbf{Reference}. While lateral velocity ($v_y$) and yaw rate performance remained consistent across configurations (Figure~\ref{fig:randomization}), longitudinal velocity ($v_x$) tracking proved highly sensitive to randomization range. The \textbf{Reference} setting achieved the lowest tracking error (RMSE: 0.2013\,m/s) compared to \textbf{Wide} (0.2651\,m/s) and \textbf{Narrow} (0.3452\,m/s), confirming that the selected parameters provide an optimal balance for this platform. The \textbf{OOD Test} range in Table~\ref{tab:domain_randomization} corresponds to the parameters used in the Widened Gap Test experiment described in the main paper.

\begin{table}[h]
\centering
\caption{Domain Randomization Range Strategies and Testing Scenarios}
\label{tab:domain_randomization}
\footnotesize
\renewcommand{\arraystretch}{1.2}
\begin{tabular}{@{}lccc@{}}
\toprule
\textbf{Parameter} & \textbf{Narrow DR} & \textbf{Wide DR} & \textbf{OOD Test} \\
\midrule
Friction coefficient & [0.8, 1.2] & [0.4, 1.6] & [0.4, 1.85] \\
Joint damping ($\mathrm{N\,m\cdot s/rad}$) & [0.0, 1.5] & [0.0, 4.0] & [0.0, 3.19] \\
Joint armature & [0.8, 1.2] & [0.4, 1.6] & [0.56, 1.44] \\
PD stiffness multiplier & [0.95, 1.05] & [0.85, 1.15] & [0.45, 1.55] \\
PD damping multiplier & [0.95, 1.05] & [0.85, 1.15] & [0.45, 1.55] \\
Torque constant & $\pm$10\% & $\pm$30\% & $\pm$22\% \\
Communication delay (ms) & [0, 5] & [0, 15] & [0, 11] \\
Link mass & [0.8, 1.2] & [0.4, 1.6] & [0.56, 1.44] \\
Push magnitude (m/s) & [0, 0.5] & [0, 0.5] & [0, 0.55] \\
Push interval (s) & 4 & 4 & 4 \\
\bottomrule
\end{tabular}
\vspace{0.05cm}
\begin{flushleft}
\scriptsize
Narrow DR: Conservative ranges for stable training. \\
Wide DR: Aggressive ranges for maximum robustness. \\
OOD Test: Out-of-distribution test extends the Reference training ranges (between Narrow and Wide) by 10\%. \\
External pushes applied at base CoM every 4 seconds with random direction.
\end{flushleft}
\end{table}

% ====================================================================
% APPENDIX E: MULTIPLE SEEDS
% ====================================================================
\section{Multiple Seed Test}
\label{app:seeds}

\begin{figure}[h]
    \centering
    \includegraphics[width=\linewidth]{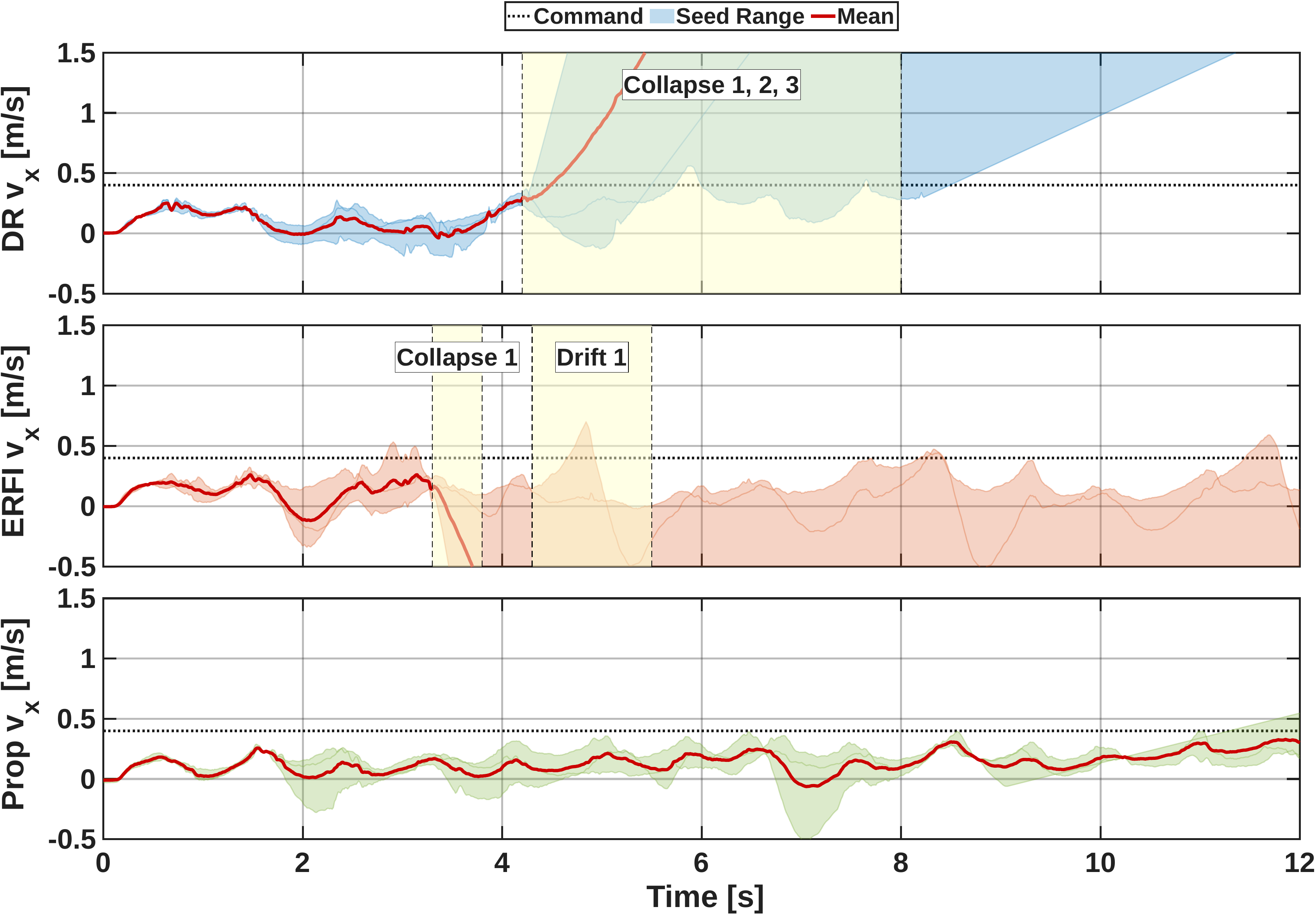}
    \caption{\textbf{Multi-Seed OOD Real-World Tracking.} Forward velocity tracking on compliant terrain with added robot mass. Shaded regions represent min-max spread across 3 seeds; red lines indicate the mean.}
    \label{fig:seed_tracking_comparison}
\end{figure}

Experiments were conducted across multiple training seeds on the physical robot. Figure~\ref{fig:seed_tracking_comparison} presents the forward velocity tracking results ($v_x$: 0.4~m/s) in OOD condition for DR, ERFI, and the Proposed method. The DR policies collapsed across all seeds. The ERFI policies demonstrate erratic behavior and tracking divergence, with inconsistent performance across seeds. The Proposed method demonstrates consistent, stable tracking across all random seeds.

Figure~\ref{fig:seed_tracking_comparison} shows the velocity tracking performance across multiple random seeds on the physical robot. The robot configuration included modified feet with contact dynamics modeling and additional batteries mounted on the base. The experiments were conducted on a compliant gym mat to evaluate performance under unseen terrain compliance.

\clearpage  % Force all pending floats to be placed before bibliography

\bibliographystyle{unsrtnat}  % Changed from plainnat/abbrvnat
\bibliography{references}

\end{document}